\pdfoutput=1
\documentclass[11pt]{article}
\usepackage{mdframed}
\usepackage[final]{acl}

\usepackage{times}
\usepackage{latexsym}

\usepackage{subfig}
\usepackage[T1]{fontenc}
\usepackage[utf8]{inputenc}

\usepackage{microtype}

\usepackage{inconsolata}

\usepackage{graphicx}

\usepackage[htt]{hyphenat}
\usepackage{algorithm}
\usepackage{algpseudocode}
\usepackage{caption}
\usepackage{float} 
\usepackage{tabularx}
\usepackage{booktabs}
\usepackage[colorinlistoftodos]{todonotes}
\usepackage{graphicx}   % For including images
\usepackage{algorithm}  % For algorithms
\usepackage{algorithmicx}
\usepackage{algpseudocode} % For pseudocode formatting
\usepackage{amsmath}  
\usepackage{tcolorbox}
\usepackage[inline]{enumitem}
\usepackage{soul}
\sethlcolor{cyan}
\usepackage{multirow}

\title{Multi-Modal Data Exploration via Language Agents}

\author{
Farhad Nooralahzadeh,
Yi Zhang,
Jonathan F{\"u}rst,
{\bf Kurt Stockinger}\\
Zurich University of Applied Sciences, Switzerland\\
 \texttt{\small\{farhad.nooralahzadeh, yi.zhang, jonathan.fuerst, kurt.stockinger\}@zhaw.ch}
}
\begin{document}
\maketitle
\begin{abstract}
International enterprises, organizations, and hospitals collect large amounts of multi-modal data stored in databases, text documents, images, and videos. While there has been recent progress in the separate fields of multi-modal data exploration as well as in database systems that automatically translate natural language questions to database query languages, the research challenge of querying both structured databases and unstructured modalities (e.g., texts, images) in natural language remains largely unexplored.
In this paper, we propose M$^2$EX~\footnote{Data and code repository are available at \url{https://github.com/yizhang-unifr/M2EX}}---a system that enables multi-modal data exploration via language agents. Our approach is based on the following research contributions: (1) Our system is inspired by a real-world use case that enables users to explore multi-modal information systems. (2) M$^2$EX leverages an LLM-based agentic AI framework to decompose a natural language question into subtasks such as text-to-SQL generation and image analysis and to orchestrate modality-specific experts in an efficient query plan. (3) Experimental results on multi-modal datasets, encompassing relational data, text, and images, demonstrate that our system outperforms state-of-the-art multi-modal exploration systems, excelling in both accuracy and various performance metrics, including query latency, API costs, and planning efficiency, thanks to the more effective utilization of the reasoning capabilities of LLMs.
\end{abstract}

\section{Introduction}

The rapid expansion of multi-modal data; spanning structured tables, text, images, and video; has created an urgent need for flexible, scalable systems for complex data exploration. In fields like healthcare, users often query across EHRs, medical images, and clinical notes using natural language. However, current systems struggle with integrating modalities, capturing user intent, and optimizing execution workflows, limiting their real-world utility.
Traditional solutions focus on single-modality tasks such as text-to-SQL~\citep{sivasubramaniamsm3,noor2024,pourreza2024din}, visual question answering~\citep{10.5555/3618408.3619222,ko-etal-2023-large,du-etal-2023-zero}, or domain-specific QA~\citep{NEURIPS2024_d0aafec0,liu-etal-2024-conversational}, often relying on rigid pipelines or handcrafted logic. While effective in narrow settings, they lack the flexibility to handle heterogeneous data or dynamic analytical goals.

Recent large language models (LLMs) and vision-language models (VLLMs) offer broader generalization but remain limited in real-world multimodal use. Techniques like retrieval-augmented generation (RAG) improve grounding but often fail at structured reasoning, long-term context, and precise tool use; especially in domains that require deep alignment across modalities and step-wise execution.
\begin{figure*}[t]
  \includegraphics[width=0.48\linewidth]{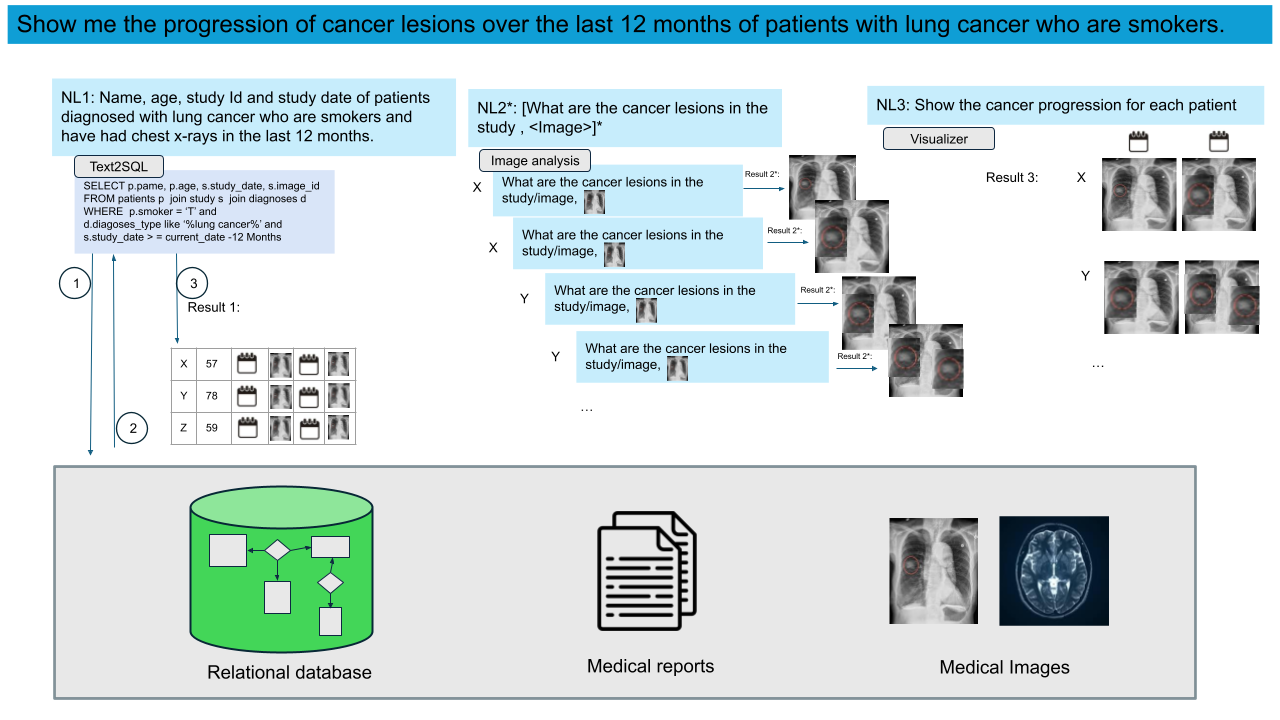} \hfill
  \includegraphics[width=0.48\linewidth]{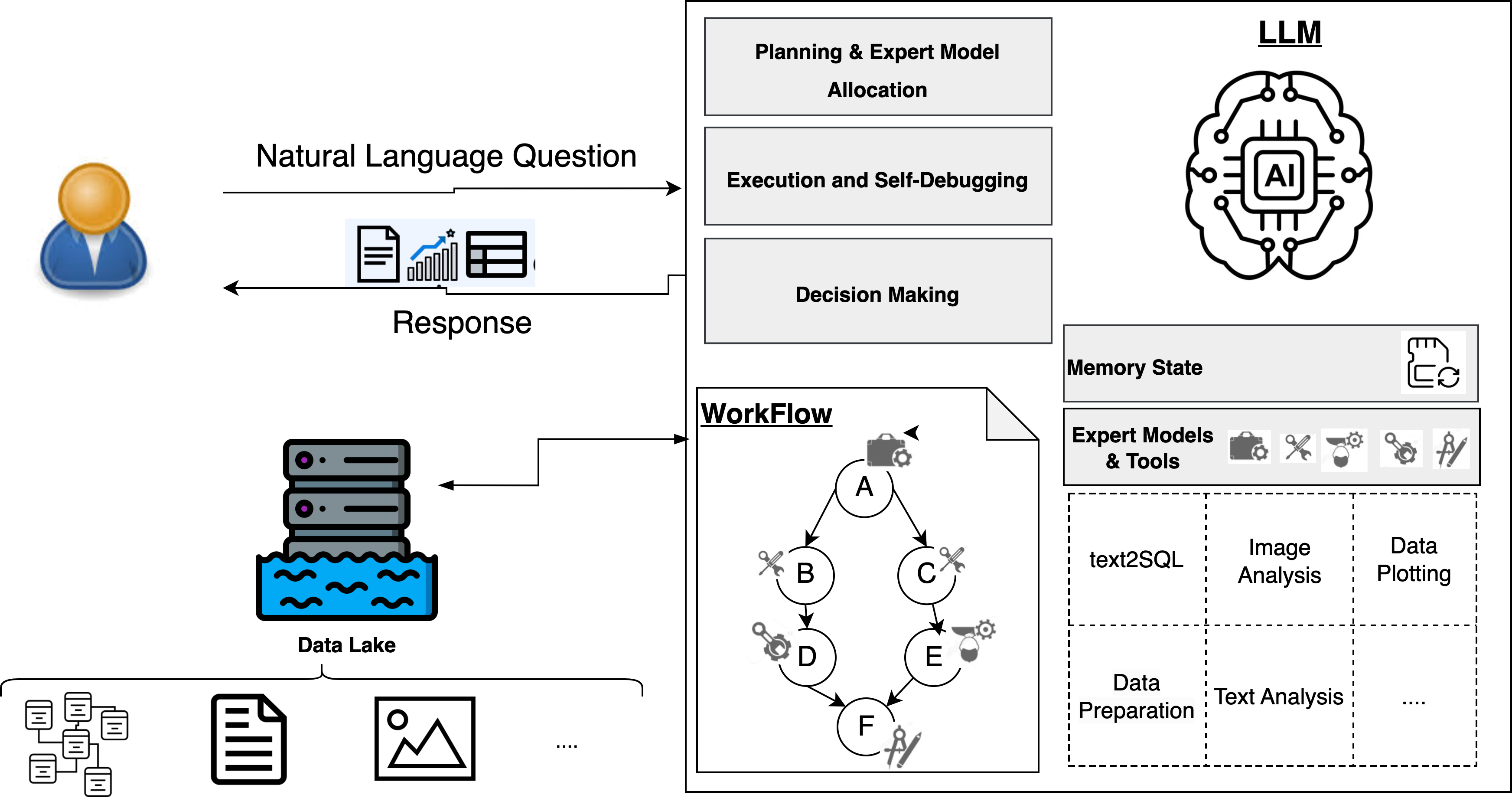}
  \caption {\textbf{(Left)}: Example workflows of multi-modal data exploration in natural language over heterogeneous data sources. \textbf{(Right)}: M$^2$EX system architecture.}
  \label{fig:XMODE_use_case}
\end{figure*}
Efforts to inspire LLMs with agentic capabilities -- such as ReAct~\citep{yao2023react}, tool invocation~\citep{yang2023mmreact,schick_toolformer_2023}, or workflow automation~\citep{liu2024declarative,urbanB24} -- have further exposed systemic challenges. 

Existing frameworks frequently adopt rigid, sequential decision-making processes, incurring computational overhead and limiting scalability. Evaluations of these systems are often conducted on in-house datasets, lacking rigorous benchmarking against ground-truth metrics or real-world multi-modal contexts. Moreover, many approaches enforce fixed task-planning hierarchies or routing mechanisms, stifling adaptability and reusability across diverse applications. This ``one-size-fits-all'' mentality contrasts starkly with the need for modular, composable agents capable of dynamically integrating domain-specific tools, retaining contextual memory, and self-optimizing workflows.

To understand these challenges, a concrete scenario of \textbf{multi-modal exploration} involving a relational database, text documents, and images is outlined here. A seemingly straightforward query like \textit{Show me the progression of cancer lesions over the last 12 months of patients with lung cancer who are smokers
} (see Figure~\ref{fig:XMODE_use_case}, Left) requires multi-modal integration, posing challenges in decomposition and optimization. 
Critical to this process is optimizing the workflow sequence, i.e., determining which queries should be executed first to minimize computational overhead and maximize efficiency.

In this work, we propose a novel framework for multi-modal data exploration that bridges these gaps through LLM-based agents designed for extensibility, precision, and cross-domain generalization. Our approach combines a ``Swiss army knife'' philosophy — enabling reusable, adaptable modules for tasks like semantic parsing, cross-modal retrieval, and structured data operations — with a principled evaluation strategy spanning diverse benchmark datasets. By decoupling task planning from execution and incorporating feedback-driven memory, our system supports iterative exploration while mitigating the pitfalls of shallow evaluation and fixed workflows. We demonstrate its efficacy across text, visual, tabular, and hybrid data domains, underscoring the potential of agentic LLMs to unify multi-modal analysis in a scalable, user-centric paradigm.

The goal of our paper is to support such multi-modal data exploration scenarios in natural language by designing and implementing a system to address the following challenges:
% \begin{itemize}[leftmargin=*, noitemsep, topsep=0pt, partopsep=0pt, parsep=0pt]

% \item \textit{Heterogeneous data exploration:} How can we design a system that accurately interprets user queries in natural language for exploring heterogeneous data sources with high accuracy?

% \item \textit{Orchestrating multiple expert models and tools for data exploration}: How can we automatically break down a user question into sub-questions that can later be organized into a workflow plan? How do we delegate these tasks to the appropriate expert models from the available toolbox, considering dependencies and the potential for parallel execution?
% \item \textit{Explainability:} How can we design a system that facilitates multi-modal exploration, allowing end users to trace conclusions back to their source data, comprehend how intermediate results were generated, and identify situations where questions remain unanswered due to missing data? 
% \end{itemize}
\begin{itemize}[leftmargin=*, noitemsep, topsep=0pt, partopsep=0pt, parsep=0pt]
\item \textit{Heterogeneous data understanding:} How can we accurately interpret natural language queries over diverse data types such as text, tables, and images?

\item \textit{Workflow orchestration:} How can we decompose a complex query into sub-tasks, organize them into an executable workflow, and delegate each to the right model or tool; respecting dependencies and enabling parallelism?

\item \textit{Explainability:} How can we provide users with traceable, transparent results, showing how answers were derived, what data contributed, and where uncertainty remains?
\end{itemize}
In this paper, we propose M$^2$EX---a multi-modal data exploration system that uses a \textit{LLM-based agentic framework} to tackle these challenges. The basic idea is to first decompose a complex natural language question into simpler sub-questions. Each sub-question is then translated into a workflow of specific tasks. By applying \textit{smart planning}, our approach can reason about which task in the workflow fails and thus re-plan that specific task rather than restarting the complete workflow. The advantage of our approach compared to similar systems such as CAESURA~\citep{urbanB24} is that it enables \textit{parallel task execution} through the construction of a directed acyclic task graph and requires a lower number of tokens from prompt engineering, resulting in more efficient query execution times and API calling costs.

The main contributions of our paper are as follows:
\begin{enumerate*}[label=(\roman*),left=0pt,labelsep=0.5em,topsep=0pt,itemsep=0pt]
% \item \textit{Higher accuracy}: M$^2$EX is based on an agentic AI framework that shows higher accuracy with improvements of up to 42\% for exploring multi-modal data than traditional work due to the smart orchestration of different tasks of the data exploration pipeline.
% \item \textit{Improved performance}: M$^2$EX demonstrates performance improvements of up to 51\% compared to state-of-the-art through parallelism, reasoning and smart re-planning
% \item \textit{Better explainability}: M$^2$EX enhances explainability by enabling a user to inspect the decisions and reasoning at each step that led to the final output, tracing back through the results of all previous steps.
% \item \textit{Generalizability}: M$^2$EX is designed and evaluated in a zero-shot setting, demonstrating its ability to perform complex tasks without relying on In-Context Learning (ICL), thereby improving both adaptability and accessibility. %See detailed \textit{Planner Prompt/Replanning Prompt} in Appendix \ref{prompts}.
\item \textbf{Unified DAG–first planning.}  
      The planner compiles a natural-language query directly into an execution \emph{directed-acyclic graph (DAG)}; independent subtasks therefore run in parallel without a second “physical-plan” stage.

\item \textbf{Self-debug \& selective re-planning for speed.}  
      Each expert tool validates its own output once; if a fault persists, the agent rewires \emph{only the affected sub-graph}.  
      This cuts end-to-end latency by up to 51\% and reduces token usage by 18\% on the ArtWork benchmark.

\item \textbf{Zero-shot cross-domain generalisation.}  
      With a single prompt set and no in-context examples, M$^{2}$EX attains up to 42\% higher answer accuracy than CAESURA and NeuralSQL on ArtWork, RotoWire, and EHRXQA.

\end{enumerate*}
\section{Related Work}
\label{sec:related_work}

\paragraph{Text-to-SQL systems.} The research field of text-to-SQL systems has seen tremendous progress over the last few years \citep{Floratou2024, pourreza2024din} due to advances in large language models. Original success can be attributed to rather simplistic datasets consisting of databases with only several tables, as in Spider \citep{yu2018spider}. Especially the introduction of new benchmarks such as ScienceBenchmark~\citep{zhang2023sciencebenchmark}, FootbalDB~\citep{furst2024evaluating}, BIRD \citep{li2024can} or SM3~\citep{sivasubramaniamsm3} has further pushed the limits of these systems. Most of the research efforts have been restricted to querying databases in English apart from a few exceptions such as Statbot.Swiss \citep{noor2024}.

\textbf{Multi-modal systems.}
Video Database Management Systems 
(VDBMSs) support efficient and complex queries over video data, but are often restricted to videos only (e.g.,~\citealp{zhang2023equi,DBLP:journals/pvldb/KangBZ19,DBLP:conf/deem/KakkarCCXVDPBS023}).  ThalamusDB~\citep{jo2024} enables queries over multi-modal data but requires SQL as input, with explicit identification of the predicates that should be applied to an attribute corresponding to video or audio data. Similarly, MindsDB\footnote{https://docs.mindsdb.com} and VIVA~\citep{DBLP:conf/cidr/KangRBKZ22} require that users write SQL and manually combine data from relational tables and models. Vision-language models provide textual descriptions of video data~\citep{zhang2024vision}, but are not designed to support precise, structured queries.
Recent multi-modal systems such as MAGMA~\cite{magma2025}, and LLaVA-Next~\cite{li2024llavanext-strong} extend vision-language reasoning via unified interfaces or tool-based controllers. However, these models are largely limited to vision-only pipelines and lack support for structured tool orchestration across modalities. In contrast, M$^2$EX generalizes to diverse tool types—including text-to-SQL, Python plotting, and image-VQA—via explicit DAG planning and partial re-planning, enabling scalable and interpretable execution across multi-modal queries.

Closest to our work are CAESURA~\citep{urbanB24}, PALIMPZEST~\citep{liu2024declarative}, and MAT~\cite{gao2025multi}, which address multi-modal querying and AI workload optimization. In contrast, M$^2$EX focuses on efficient orchestration of model calls and dependencies, reducing latency and cost while improving accuracy by minimizing interference from intermediate outputs~\citep{schick_toolformer_2023}\footnote{CAESURA and MAT employ the ReAct agent framework, which leads to extended context tokens and increased latency.}.

While related systems emphasize query planning, they fall short in enhancing the \textit{accuracy} and \textit{explainability} of model outputs—critical needs in domains like medical data science, where regulatory standards require transparent and justifiable results.

\section{Method and System Design}
\paragraph{Problem statement.} Given a multi-modal query $q$, a data lake $D$, a tool catalogue $T$ with metadata $T_{\text{meta}}$, our goal is to produce a directed acyclic task graph
$G=(V,E)$ and a final answer $a$ such that
each node $v\!\in\!V$ is a \textit{(tool,\,args)} pair,
edges $E$ encode data dependencies,
the execution of $G$ is valid w.r.t.\ $T_{\text{meta}}$,
and $a$ maximizes task-level answer accuracy.
\paragraph{Proposed System:} To address this problem, M$^2$EX enables \underline{m}ulti-\underline{m}odal data \underline{ex}ploration via language agents. Its details are presented in Algorithm~\ref{alg:workflow} and Figure \ref{fig:XMODE_use_case} (right) (A fully annotated DAG and end-to-end use-case example appear in Figures \ref{fig:XMODE-ehr} and \ref{fig:XMODE-artwork}). 
M$^2$EX is an \textit{agentic system}~\citep{kapoor2024ai} driven by LLMCompiler~\citep{kim2023llm}, a dynamic planner pattern based on a Large Language Model, equipped with a comprehensive toolkit \( \mathcal{T} \) containing all the necessary models to decompose a user’s request, such as a multi-modal natural language question, into a workflow (i.e., a graph of sub-questions). The workflow is represented as a Directed Acyclic Graph (\textit{DAG}), where each node corresponds to a simple sub-task (or sub-question) with a specific tool assigned by the planner. While decoupling logical and physical plans can be suboptimal due to plan ambiguity and nonlinearity, unlike CAESURA, the planner determines sub-tasks that can be executed in parallel and manages their dependencies by leveraging an LLM to directly generate the execution plan from the query as a graph of function calls. M$^2$EX is designed to be adaptable, allowing dynamic debugging and plan modification (re-planning) when necessary, for example, if a failure occurs during a text-to-SQL sub-task.
\begin{algorithm}[t]
\caption{M$^2$EX: Multi-Modal Data Exploration via Language Agents}
\label{alg:workflow}
\begin{algorithmic}[1]
\scriptsize%% Reduce font size further
\renewcommand{\alglinenumber}[1]{\scriptsize#1} % Smaller line numbers
\setlength{\textfloatsep}{5pt} % Reduce space around algorithm
\setlength{\floatsep}{5pt}

\Require User query \( q \), Agent Core \( \mathcal{LLM} \), toolkit \( \mathcal{T} \), Data Lake \( D \), Pre-defined Prompts \( \mathcal{P} \), Empty memory state \( \mathcal{R} \)
\Ensure Final answer \( a \)

\State \textbf{Stage 1: Planning \& Expert Model Allocation}
 \State \( \mathcal{R} \gets \mathcal{R} \cup \{q, D_{meta}\} \)
\State \( S \gets \textsc{decompose}( \mathcal{R}, \mathcal{LLM},\  \mathcal{T}_{meta} )  \)
\Comment{Use an agent core \( \mathcal{LLM} \) (with a planner prompt \( \in \mathcal{P} \) access to tool metadata) to decompose \( q \) into subtasks \( s_1,\dots,s_n \). Each task contains a tool, arguments, and list of dependencies.}

\State \( G \gets \textsc{BuildDAG}(\mathcal{S},\mathcal{LLM}) \) 
\Comment{Construct a Directed Acyclic Graph (DAG): \( G \) where each node represents a subtask and edges represent dependencies}
 
\State \textbf{Stage 2: Execution \& Self-debugging}
\State \( \sigma \gets\textsc{TopologicalSort}(G) \) 
\Comment{Determine an execution order that respects dependencies}
\State \( \mathcal{B} \gets \textsc{GroupParallelTasks}(\sigma, G) \) 
\Comment{Partition tasks into parallel execution}
\For{each batch  \( b_k \in \mathcal{B} \)}
    \State Launch parallel execution:
    \For{each subtask \( s_i \in b_k \)}
        \State \( r_i \gets \textsc{Execute}(s_i, \mathcal{T} ,\mathcal{D} )  \) 
        \Comment{Invoke the assigned expert tool for \( s_i \). Integrate $n$-time self-debugging to automatically detect and correct errors as needed. ($n=1$). If there is still an error, provide an error message as an output of execution. }
        \State \( \mathcal{R} \gets \mathcal{R} \cup \{r_i\} \)
    \EndFor
\EndFor
    \State \textbf{Stage 3: Decision Making}

\State Validate \( \mathcal{R} \) via reflection 
    \Comment{Check that outputs are correct and executable; if not, trigger error feedback.}
    \If{validation fails}
        \State \( G \gets \textsc{Replan}(G, \mathcal{R},\mathcal{LLM},\  \mathcal{T}_{meta}) \)
        \Comment{Dynamically adjust the DAG (e.g., reallocate tasks or update tool parameters) based on error feedback using an agent core \( \mathcal{LLM} \) (with a replanning prompt \( \in \mathcal{P} \)).}
        \State \textbf{goto} line 5 
        \Comment{Restart execution with the updated plan.}
    \EndIf
\State \( a \gets \textsc{Synthesize}(\mathcal{R}, \mathcal{LLM}) \) 
\Comment{Aggregate and refine intermediate results into the final answer using LLM reasoning.}
\If{\( a \) is insufficient or uncertain}
    \State \( G \gets \textsc{Replan}(G, \mathcal{R},\mathcal{LLM},\  \mathcal{T}_{meta}) \)
        \Comment{Dynamically adjust the DAG (e.g., reallocate tasks or update tool parameters) based on error feedback  using an agent core \( \mathcal{LLM} \) (with a replanning prompt \( \in \mathcal{P} \)).}
        \State \textbf{goto} line 5
        \Comment{Restart execution with the updated plan.}
\EndIf
\State \Return \( a \)
\end{algorithmic}
\end{algorithm}

As shown in Algorithm~\ref{alg:workflow} and Figure \ref{fig:XMODE_use_case}, the system is composed of the following key components:
\begin{enumerate*}[label=(\arabic*),left=0pt,labelsep=0.5em,topsep=0pt,itemsep=0pt]
    \item \textit{User Query (\( q \)):} a multi-modal natural language question posed by the user, which initiates the process of task decomposition and execution.
    \item \textit{Agent Core (\( \mathcal{LLM} \)):} the core reasoning engine that powers the dynamic planning, execution, and decision-making processes. The LLM is responsible for decomposing the user query into subtasks, managing dependencies, and synthesizing final results using diverse prompts~\( \mathcal{P}\).
    \item \textit{Expert Models \& Tools ( Toolkit ) (\( \mathcal{T} \)):} a comprehensive collection of expert models and tools that are used for executing specific sub-tasks. The toolkit provides the necessary models for tasks such as \texttt{text-to-SQL}, \texttt{text analysis}, \texttt{image analysis}, \texttt{data preparation}, and \texttt{data plotting}.  Each expert model or tool should include a description and argument specifications (\( \mathcal{T}_{meta} \)), and they will be available during the planning and re-planning stages.
    \item \textit{Data Lake (\( D \)):} a central repository that stores both structured and unstructured data, such as tabular data, images, and text. Each expert model and tool has direct access to the data lake to perform its assigned tasks. The data stored in the lake is utilized as input for various tasks, enabling the system to generate accurate results for the user’s query.
    \item \textit{Pre-defined Prompts (\( \mathcal{P} \)):} a collection of predefined prompts available to the {LLM}, which are used to guide the reasoning process during planning, execution, and decision-making (see details in Appendix \ref{prompts}).
    \item \textit{Memory State (\( \mathcal{R} \)):} The initial memory state starts empty and captures all intermediate results and interactions throughout the workflow execution. The system tracks these intermediate results using an output object that stores the answer and reasoning at each node in the workflow.
    \item \textit{Final Answer (\( a \)):} The final answer is the output generated by the system after executing all the tasks and performing reasoning through the LLM. It consolidates all intermediate results and provides a comprehensive response to the user’s query. 
    
    The final answer typically includes several components: a summary of the task or query result, detailed information about the outcome, the source of the data used, an inference indicating the success of the task, and any additional explanations or clarifications. This structured output ensures that the user receives not only the result but also the reasoning and context behind it. 
\end{enumerate*}
In Figure ~\ref{fig:XMODE-ehr}, we demonstrate the showcase of M$^2$EX using an example query applied to the EHRXQA data, which includes relational tables and images: \textit{Was patient 18061894 prescribed acetaminophen, and did a chest x-ray show any technical assessments until 12/2103?}

The system starts with the user query \( q \) and processes it through several stages, as detailed below:
\begin{enumerate}[label=(\roman*), 
                  left=0pt, 
                  labelsep=0.3em,   % Decrease horizontal space between label and item
                  topsep=0pt, 
                  partopsep=0pt,    % Prevent extra space above if nested
                  itemsep=0pt, 
                  parsep=0pt,       % Remove paragraph spacing within items
                  wide=0pt]
    \item \textit{Planning \& Expert Model Allocation.} 
    The system begins by analyzing the user query \( q \) and decomposes it into a sequence of tasks. Using the agent core (\( \mathcal{LLM} \)), the system identifies the required expert models and tools from the toolkit \( \mathcal{T} \), along with their input arguments and inter-dependencies. These subtasks are synthesized into a workflow represented as a Directed Acyclic Graph (\textit{DAG}), \( G \), where each node represents a task, and edges represent dependencies between them. E.g., a natural language question can be split into multiple tasks such as \texttt{intent table detection}, \texttt{text2SQL}, and \texttt{image analysis} as shown in Figure~\ref{fig:XMODE-ehr}. The workflow reflects the execution sequence and dependencies that are necessary to answer the user’s query. The system also utilizes predefined prompts \( \mathcal{P} \) to guide the reasoning process during task decomposition.
    \item \textit{Execution and Self-Debugging.} 
    The system executes the tasks according to the generated workflow by invoking the relevant expert models and tools from the toolkit \( \mathcal{T} \). The system utilizes a state object \( \mathcal{R} \), which stores intermediate results and interactions during the execution. The tasks are partitioned into independent batches \( \mathcal{B} \) that can be executed in parallel, which is determined through a topological sort (\textsc{TopologicalSort}(\( G \))) of the DAG. For each batch, the system launches parallel executions of the assigned tasks. The tasks are executed using the expert models, and the outcomes are passed on to subsequent tasks that depend on them. Each expert model includes a self-debugging mechanism to detect and correct errors during execution. If an error persists, the system can provide feedback and retry the process, thereby enhancing the robustness of the execution.
    \item \textit{Decision Making.} After the execution of the subtasks, M$^2$EX inspects the intermediate results stored in \( \mathcal{R} \) to determine whether they are sufficient to fulfill the user’s request. If the results are satisfactory, the system synthesizes them into the final answer \( a \). However, if the results are insufficient or uncertain, the system triggers a re-planning process by invoking \textsc{Replan}(\( G, \mathcal{R}, \mathcal{LLM}, \mathcal{T}_{\text{meta}} \)) to adjust the DAG and re-execute the tasks. This process repeats until the decision-making component is satisfied with the final result or a predefined maximum loop limit is reached.
\end{enumerate}
In summary, M$^2$EX uses an algorithmic approach where the system first decomposes the user query into subtasks, executes these tasks with error detection and correction mechanisms, and synthesizes the results into a final answer. The system is highly adaptive, with dynamic re-planning capabilities powered by the reasoning abilities of the {LLM} to ensure efficient task execution, debugging, and modification of the plan when needed.
Our current M$^2$EX implementation offers a range of features, including self-debugging, query re-planning, optimization, and explainability to better understand how a natural language question is decomposed into multiple sub-tasks. See details in Appendix \ref{opts}.
\paragraph{Complexity and convergence.} Planning inspects $|S|$ subtasks and calls $\Phi$ once, costing $\mathcal{O}(|S|\,C_{\text{LLM}})$ tokens
($C_{\text{LLM}}$ = context length processed by the language model). With unlimited workers, execution latency is $\mathcal{O}(\text{depth}(G))$; with $p$ workers it is bounded by $\text{depth}(G)$ as well. Re-planning only touches the affected sub-DAG, so its worst-case cost is strictly $\leq$ the first planning pass.
\section{Experiments}

In this section, we evaluate M$^2$EX's performance, focusing on the following research questions:
% \begin{enumerate*}[label=(\arabic*),left=0pt,labelsep=0.5em,topsep=0pt,itemsep=0pt]
% \begin{itemize}[leftmargin=*,noitemsep,topsep=0pt]
  % \setlength\itemsep{1em}
(1) How well does M$^2$EX tackle multi-modal natural language questions on three different datasets consisting of tabular data and images?
(2) How does the system perform compared to state-of-the-art systems  such as CAESURA \citep{urbanB24} and NeuralSQL \citep{bae2024ehrxqa} on underlying benchmark datasets?
(3) What systematic errors can we observe?
% \end{enumerate*}
\subsection{Experimental Setup}
\textbf{Datasets} For our experiments, we used three different datasets, namely datasets about artwork, basketball, as well as electronic health records. Due to hardware limitations, we reduced the dataset to 100 images and reports. Processing the full size in CAESURA can result in crashes due to out-of-memory issues. 
\begin{enumerate}[label= \textsc{Dataset \arabic*:} , 
                  left=0pt, 
                  labelsep=0.3em,   % Decrease horizontal space between label and item
                  topsep=0pt, 
                  partopsep=0pt,    % Prevent extra space above if nested
                  itemsep=0pt, 
                  parsep=0pt,       % Remove paragraph spacing within items
                  wide=0pt]
\item \textsc{ArtWork.} This dataset was introduced by \citet{urbanB24} and contains information about paintings in tabular form as well as an image collection containing 100 images of the artworks, collected from Wikipedia. 
The tabular data contains metadata about paintings such as title,  inception, movement, etc. as well as a reference to the respective paintings. A typical example question from this dataset is \textit{Plot the number of
paintings depicting war for each century} (see Figure \ref{fig:XMODE-artwork} in the Appendix). 

In addition to the 24 existing questions in the ArtWork dataset, we propose six new questions aimed at evaluating parallel task planning and execution, facilitating a comparison between the characteristics of the two architectures. These six questions incorporate both single and multiple modalities. Moreover, four of the six questions require responses in various formats: two questions demand two plots, and two questions involve a combination of plotting and showing the results in a specific data structure, i.e. either as a tabular format or as a JSON format.
The final test dataset contains 30 natural language questions derived from the original 24 in the ArtWork dataset. These include 8 queries seeking a single result value, 11 requiring structured data as output, and 11 requesting a plot. Of these, 18 queries involve multi-modal data, while the remaining 12 are based exclusively on relational data.
We have chosen this dataset to directly compare our system with CAESURA~\citep{urbanB24}, one of the state-of-the-art systems for multi-modal data exploration in natural language.
\begin{figure*}[t]
    \centering
\includegraphics[width=1\linewidth]{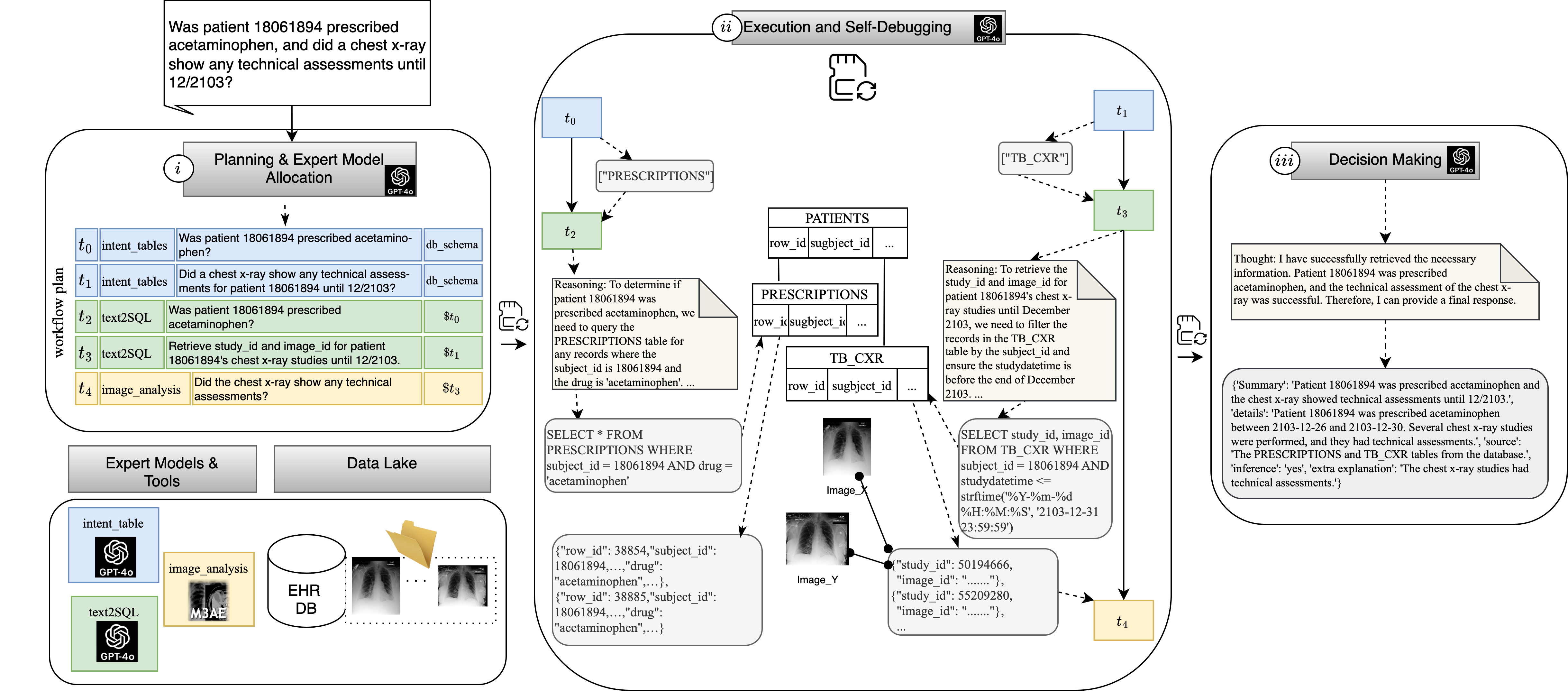}
\caption{M$^2$EX system architecture in EHRXQA ~\citep{bae2024ehrxqa} with an example of processing a multi-modal query. The query is automatically decomposed into various components which can be inspected by the user for explainability.}
\label{fig:XMODE-ehr}
\end{figure*}
\item \textsc{RotoWire.} 
This dataset is also utilized by \citet{urbanB24} and consists of one relational database and 100 randomly selected textual reports about NBA games, including metadata, key statistics of individual players, and team performance metrics. A typical example question from this dataset is \textit{Plot the highest number of three-pointers made by players from each nationality}.
The test dataset comprises 12 natural language questions, evenly divided into 6 single-modal and 6 multi-modal queries. Regarding output format, 3 questions require a single value as a response, 5 involve structured data outputs, and 4 necessitate visualization through plots. 

\item \textsc{Electronic Health Records (EHR).} We also utilized the EHRXQA~\citep{bae2024ehrxqa} dataset, a multi-modal question answering dataset that integrates structured electronic health records (EHRs) with chest X-ray images. This dataset consists of 18 tables and 432 images, and specifically requires cross-modal reasoning. The questions of EHRXQA are categorized based on their scope in terms of modality and patient relevance.
For \textit{modality-based} categorization, questions were classified into three types: Table-related, image-related,  and table-image-related, based on the data modality required. The \textit{patient-based} categorization classified questions based on their relevance to a single patient, a group of patients, or none (i.e., unrelated to specific patients).
We have chosen this dataset since it was used to evaluate NeuralSQL, another state-of-the-art system for multi-modal data exploration. To manage the cost of an API call, we extracted 100 questions randomly. The selection process was guided by three predefined categories within the test set of the EHRXQA dataset: Image Single-1, Image Single-2, and Image+Table Single (for details, please look at \citet{bae2024ehrxqa}).
\end{enumerate}
Several considerations influenced our decision to work with reduced versions of these datasets: 
\begin{enumerate*}[label={}, itemjoin={{ }},afterlabel={}]
\item {\textit{Demonstrating Viability} The reduced dataset size demonstrates M$^2$EX's viability across diverse multi-modal datasets with ground truth, proving its ability to handle complex queries in a controlled setting.} 
\item{\textit{Complexity of Building Datasets} Constructing large-scale multi-modal datasets with precise ground truth is a complex, manual process, which limits the scaling-up within the study's scope.}
\item{\textit{Cost Considerations} The cost of API calls to the LLM powering M$^2$EX necessitates a balance between dataset size and experimental feasibility, ensuring thorough evaluation within practical constraints.}
\end{enumerate*}

\subsection{Baseline Systems and Setup} 

We compare M$^2$EX to the baseline implementations of CAESURA \citep{urbanB24}  and  NeuralSQL \citep{bae2024ehrxqa} - two important state-of-the-art systems for multi-modal data exploration. 

CAESURA supports natural language queries over a multi-modal data lake, leveraging
BLIP-2 \citep{li2023blip} for visual question answering and a fine-tuned BART \citep{lewis2020bart} for text question answering.
We reproduced the results of CAESURA on the ArtWork and RotoWire datasets using GPT-4o for planning, data processing, and plot generation while adopting the other tool models as proposed in CAESURA \citep{urbanB24}. For comparison with our system, we use GPT-4o as the LLM for both planning and text analysis on RotoWire. On ArtWork, we employ GPT-4o as the planner and retain the same model for visual question answering (i.e., BLIP-2) in M$^2$EX.

In NeuralSQL, an LLM is integrated with an external visual question answering system, M3AE model~\citep{10.1007/978-3-031-16443-9_65}, to handle multi-modal questions over a structured database with images by translating a user question to SQL in one step.  
To ensure that we used the optimal hyperparameter settings and prompt structure, we contacted the authors of EHRXQA~\citep{bae2024ehrxqa}, who provided the results of their experiment for NeuralSQL using GPT-4o on 100 randomly selected questions.

For M$^2$EX, we employ the M3AE model with task-specific fine-tuned weights, provided by ~\citep{bae2024ehrxqa}, for the image analysis task. The customized M3AE model is encapsulated as a web service and deployed on the same computing node as our experiments.
We conduct the experiments using a CUDA-accelerated computational node on an OpenStack virtual host. This node is equipped with a 16-core CPU, 16 GB of main memory, and 240 GB of SSD storage. Additionally, it features an NVIDIA T4 GPU with 16 GB of dedicated graphics memory.
A complete mapping of subtasks to expert models/tools and prompt types is provided in Table~\ref{tab:tool-table} (Appendix~\ref{sec:appendix-tools}).
\subsection{Evaluation Metrics}
\label{sec:metrics}
To evaluate M$^2$EX against state-of-the-art systems, we use the following metrics:
\begin{enumerate*}[label=(\roman*)]
% \begin{itemize}[leftmargin=*,noitemsep,topsep=2pt]
\item \textit{Accuracy}: Measures the accuracy (i.e., exact match) of the generated result set compared with the gold standard result set or with the human expert.
\item \textit{Steps}: Number of steps required by the respective system to come up with the final result. These steps include reasoning, planning, re-planning, etc.
\item \textit{Tokens}: Number of tokens used for prompt engineering. 
\item \textit{Latency}: End-to-end execution time for a system to come up with the final result.
\item \textit{API costs}: Costs for calling the LLM, e.g. for GPT4o.
\end{enumerate*}

We apply the above-mentioned metrics under various questions and system categories:

% \begin{itemize}[leftmargin=*,noitemsep,topsep=2pt]
\begin{enumerate*}[label=(\roman*)]
\item \textit{Modality}: Questions can either be of \textit{single} modality, i.e., querying only relational data or image data, or of \textit{multiple} modalities, i.e., querying both relational and image data.
\item \textit{Output Type}: The output type of a question can either be a \textit{single value}, e.g., true or false, a \textit{data structure}, e.g., in tabular or JSON format, a plot, or a combination of plots and data structures.
\item \textit{Workflow}: The generated workflow plan can either be \textit{sequential} or \textit{parallel}.
\end{enumerate*} Finally, we evaluate if a system generates a correct (multi-modal) query plan (i.e., generated plan), and if it supports re-planning.

%\subsection{Hardware Setup}
%\label{section: hardware-setup}

%\begin{itemize}
%    \item Evaluating XMODE on both Artwork and EHRXQA datasets  
%    \item Evaluating CAESURA on the Artwork dataset.
%\end{itemize}
\subsection{Results on the Benchmark Datasets}\label{sec:Benchmark} 

%% end of block comment

\begin{table*}[t]
\setlength\tabcolsep{2pt}
    \centering
    \resizebox{.95\textwidth}{!}{
\begin{tabular}{@{}l|ll|r|r|r|r|c|c|r|r|r|r|r|c|c@{}} %% 3+6+6+1
\toprule
\multirow{2}{*}{\textbf{System}}
& \multicolumn{2}{|c|}{\multirow{2}{*}{\textbf{Category} (\# in ArtWork$|$\# in RotoWire)}} & \multicolumn{6}{c|}{\textbf{ArtWork}} & \multicolumn{6}{c|}{\textbf{RotoWire}} & \multirow{2}{*}{\textbf{Re-planning}}\\
\cmidrule(lr){4-9} \cmidrule(lr){10-15}
& & & \textbf{Accuracy} & \textbf{Steps} & \textbf{Tokens} & 
\textbf{Latency [s]} & \textbf{Cost [\$]} & 
\textbf{Gen. Plan} & \textbf{Accuracy} & \textbf{Steps} & \textbf{Tokens} & 
\textbf{Latency [s]} & \textbf{Cost [\$]} & 
\textbf{Gen. Plan} &  \\ 
\midrule
 \multirow{10}{*}{\textbf{CAESURA}} & 
 \multicolumn{1}{l|}{\multirow{2}{*}{\textbf{Modality}}} & Single ($15|6$) & 60.00\% & 152 & 214,014 & 973.28 & 1.33 & 
 \multirow{10}{*}{80\%} & 50.00\% & 79 & 100,277 & 500.52 & 0.65 & \multirow{10}{*}{91.67\%} & \multirow{10}{*}{No} \\
 & \multicolumn{1}{l|}{} & Multiple ($15|6$) & 6.67\% & 164 & 268,918 & 4,847.95 & 1.65 & 
 & 0.00\% & 78 & 133,230 & 959.17 & 0.85 & \\ \cmidrule(lr){2-8} \cmidrule(lr){10-14}
 & \multicolumn{1}{l|}{\multirow{5}{*}{\textbf{Output}}} & Single Value ($8|3$) & 37.50\% & 88 & 135,077 & 1,047.24 & 0.82 &  
 & 66.67\% & 32 & 45,145 & 287.55 & 0.29 & \\ 
 & \multicolumn{1}{l|}{} & Data Structure ($10|5$) & 50.00\% & 116 & 183,454 & 2,683.03 & 1.14 & 
 & 20.00\% & 69 & 104,345 & 659.37 & 0.68 & & \\ 
 & \multicolumn{1}{l|}{} & Plot ($8|4$) & 25.00\% & 79 & 112,732 & 1,856.66 & 0.69 & 
 & 0.00\% & 56 & 84,017 & 512.77 & 0.53 & &\\ 
 \small{few-shot ($4$)}  & \multicolumn{1}{l|}{\textbf{Type}} & Plot-Plot ($2|0$) & 0\% & 16 & 21,508 & 108.87 & 0.14 & 
 & -- & -- & -- & -- & -- & &  \\ 
\small{in planning} & \multicolumn{1}{l|}{} & Plot-Data Structure ($2|0$) & 0\% & 17 & 30,161 & 125.42 & 0.19 & 
 & -- & -- & -- & -- & -- & &  \\ \cmidrule(lr){2-8} \cmidrule(lr){10-14}
 & \multicolumn{1}{l|}{\multirow{2}{*}{\textbf{Workflow}}} & Sequential ($24|12$) & 41.67\% & 261 & 399,045 & 5,330.12 & 2.45 & 
 & 25.00\% & 157 & 233,507 & 1,459.69 & 1.50 &  & \\ 
 & \multicolumn{1}{l|}{} & Parallel ($6|0$) & 0\% & 55 & 83,887 & 491.11 & 0.52 &
 & -- & -- & -- & -- & -- & &  \\ \cmidrule(lr){2-8} \cmidrule(lr){10-14}
 & \multicolumn{2}{c|}{\textbf{Overall} ($30|12$)} & 33.33\% & 316 & \textbf{482,932} & 5,821.23 & 2.98 & 
 & 25.00\% & 157 & \textbf{233,507} & \textbf{1,459.69} & \textbf{1.50} &  &  \\ 
 
\midrule
\multirow{10}{*}{\textbf{M$^2$EX}} & 
 \multicolumn{1}{l|}{\multirow{2}{*}{\textbf{Modality}}} & Single ($15|6$) & 100.00\% & 96 & 159,212 & 525.09 & 0.61 & 
 & 100.00\% & 34 & 89,810 & 524.06 & 0.40 & & \\
 & \multicolumn{1}{l|}{} & Multiple ($15|6$) & 26.67\% & 107 & 326,400 & 2,515.03 & 1.49 & \multirow{10}{*}{100\%} 
 &  33.33\% & 42 & 952,386 & 3,235.96 & 3.22 & \multirow{10}{*}{100\%} &  \multirow{10}{*}{Yes}  \\ 
 \cmidrule(lr){2-8} \cmidrule(lr){10-14}
& \multicolumn{1}{l|}{\multirow{5}{*}{\textbf{Output}}} & Single Value ($8|3$) & 50.00\% & 56 & 71,575 & 494.78 & 0.39 & 
& 100.00\% & 16 & 108,520 & 499.70 & 0.40 &  & \\ 
 & \multicolumn{1}{l|}{} & Data Structure ($10|5$) & 50.00\% & 67 & 223,528 & 1,330.40 & 0.89 & 
 & 40.00\% & 27 & 410,698 & 2,120.15 & 1.57 & & \\ 
 & \multicolumn{1}{l|}{} & Plot ($8|4$) & 75.00\% & 52 & 118,431 & 798.97 & 0.48 & 
 & 75.00\% & 33 & 522,987 & 1,140.17 & 1.65 & & \\ 
 \small{zero-shot}& \multicolumn{1}{l|}{\textbf{Type}} & Plot-Plot ($2|0$) & 100.00\% & 14 & 50,108 & 308.92 & 0.22 & 
 & -- & -- & -- & -- & -- & & \\ 
 & \multicolumn{1}{l|}{} & Plot-Data Structure ($2|0$) & 100.00\% & 14 & 21,970 & 107.05 & 0.10 & 
 & -- & -- & -- & -- & -- & &  \\  \cmidrule(lr){2-8} \cmidrule(lr){10-14}
 & \multicolumn{1}{l|}{\multirow{2}{*}{\textbf{Workflow}}} & Sequential ($24|12$) & 62.50\% & 163 & 338,766 & 2,131.11 & 1.51 & 
 & 66.67\% & 76 & 1,042,196 & 3,760.02 & 3.62 &  & \\ 
 & \multicolumn{1}{l|}{} & Parallel ($6|0$) & 66.67\% & 40 & 146,846 & 909.01 & 0.59 & 
 & -- & -- & -- & -- & -- & &  \\ \cmidrule(lr){2-8} \cmidrule(lr){10-14}
 & \multicolumn{2}{c|}{\textbf{Overall} ($30|12$)} & \textbf{63.33\%} & \textbf{203} & 485,612 & \textbf{3,040.12} & \textbf{2.10} & 
 & \textbf{66.67\%} & \textbf{76} & 1,042,196 & 3,760.02 & 3.62 &  &  \\ 
\bottomrule
\end{tabular}
}

\caption{Performance metrics of Caesura \citep{urbanB24} and M$^2$EX on ArtWork and RotoWire. Planner coverage (\textit{Gen.\ Plan}) is 100\% on ArtWork and RotoWire, indicating reliable task decomposition across domains.}
\label{tab:artwork_rotowire}
\end{table*}

% \begin{table*}[t]
% \setlength\tabcolsep{1pt}
%     \centering
%     \resizebox{\textwidth}{!}{
% \begin{tabular}{@{}lc|r|r|r|r|r|r|c|c@{}}
% \toprule
% \multicolumn{2}{c}{\multirow{3}{*}{\textbf{System}}} &  \multicolumn{3}{|c|}{\textbf{Scope}} & \multicolumn{2}{c|}{\textbf{Output Type}}   & \multirow{3}{*}{\textbf{Overall (100)} }&{\textbf{Generated}}  & \multirow{3}{*}{\textbf{Replanning}}\\
% \cmidrule(lr){3-5} \cmidrule(lr){6-7}
%  & &Image Single-1 & Image Single-2  & Image+Table Single  & Binary & Categorical & & \multirow{2}{*}{\textbf{Plan}}&   \\
%  & &(30)& (30) &(40)  & (50) & (50) &  & &  \\
% \midrule

% \multirow{3}{*}{\textbf{NeuralSQL}} &\small{zero-shot}& 0.00\% & 0.00\% & 0.00\% & 0.00\% & 0.00\% & 0.00\%& \multirow{3}{*}{N/A} &   \multirow{3}{*}{No} \\

% & &  &   &   &   &  &   &  &    \\
% & \small{few-shot} \tiny{($n=10$)} & \textbf{26.67}\% & 20.00\% & 47.50\% & 48.00\% & 18.00\% & 33.00\% &      \\
% \midrule
% \textbf{M$^2$EX} &\small{zero-shot}    & 23.33\% & \textbf{43.33}\% & \textbf{77.50}\% & \textbf{74.00}\% & \textbf{28.00}\% & \textbf{51.00\%} & \textbf{98}\% &  \textbf{Yes}  \\
% \bottomrule
% \end{tabular}
% }

% \caption{Performance metrics of NeuralSQL (zero-shot and few-shot) and M$^2$EX (zero-shot) on EHRXQA. Planner coverage (\textit{Generated Plan}) is 98\% on EHRXQA, indicating reliable task decomposition across domains.}
% \label{tab:ehr}
% %\vspace{-5pt}
% \end{table*}

\begin{table*}[t]
  \centering
  %%% First table: ~0.68\textwidth %%%
  \begin{minipage}[t]{0.68\textwidth}
    \small
\setlength\tabcolsep{1pt}
    \resizebox{\textwidth}{!}{%
      \begin{tabular}{@{}lc|r|r|r|r|r|r|c|c@{}}
        \toprule
        \multicolumn{2}{c}{\multirow{3}{*}{\textbf{System}}} 
          &  \multicolumn{3}{|c|}{\textbf{Scope}} 
          & \multicolumn{2}{c|}{\textbf{Output Type}}   
          & \multirow{3}{*}{\textbf{Overall (100)}} 
          & {\textbf{Generated}}  
          & \multirow{3}{*}{\textbf{Replanning}}\\
        \cmidrule(lr){3-5} \cmidrule(lr){6-7}
         & &Image Single-1 & Image Single-2  & Image+Table Single  
            & Binary & Categorical & & \multirow{2}{*}{\textbf{Plan}}&   \\
         & &(30)& (30) &(40)  & (50) & (50) &  & &  \\
        \midrule

        \multirow{3}{*}{\textbf{NeuralSQL}} 
          &\small{zero-shot}& 0.00\% & 0.00\% & 0.00\% 
            & 0.00\% & 0.00\% & 0.00\% & \multirow{3}{*}{N/A} 
            & \multirow{3}{*}{No} \\
        & &  &   &   &   &  &   &  &    \\
        & \small{few-shot} \tiny{($n=10$)} 
            & \textbf{26.67}\% & 20.00\% & 47.50\% 
            & 48.00\% & 18.00\% & 33.00\% &      \\
        \midrule
        \textbf{M$^2$EX} &\small{zero-shot}    
          & 23.33\% & \textbf{43.33}\% & \textbf{77.50}\% 
          & \textbf{74.00}\% & \textbf{28.00}\% & \textbf{51.00\%} 
          & \textbf{98}\% &  \textbf{Yes}  \\
        \bottomrule
      \end{tabular}%
    }
    \caption{Performance metrics of NeuralSQL (zero-shot and few-shot) and M$^2$EX (zero-shot) on EHRXQA. Planner coverage (\emph{Generated Plan}): 98\%.}
    \label{tab:ehr}
  \end{minipage}\hfill
  %%% Second table: ~0.27\textwidth %%%
  \begin{minipage}[t]{0.3\textwidth}
    \scriptsize
    \setlength{\tabcolsep}{3pt}
    \resizebox{\textwidth}{!}{%
      \begin{tabular}{@{}lcccl@{}}
        \toprule
        \textbf{Dataset} & \textbf{System} & \textbf{Tasks} 
          & \textbf{Errors} & \textbf{Dominant Error Source} \\
        \midrule
        ArtWork  & CAESURA & 30 & 20 & Faulty plans; VQA errors \\
                 & M$^2$EX & 30 & 11 & VQA errors only \\
        RotoWire & CAESURA & 12 &  9 & Text analysis; SQL faults \\
                 & M$^2$EX & 12 &  4 & Text analysis \\
        EHRXQA   & NeuralSQL&100&67& N/A – no plan output\\
                 & M$^2$EX  &100 & 49 & VQA errors only\\
        \bottomrule
      \end{tabular}%
    }
    \caption{Top-level error breakdown. See App.~\ref{err} for details.}
    \label{tab:error-headline}
  \end{minipage}
\end{table*}

\paragraph{Results on the ArtWork and RotoWire Datasets} 

Table \ref{tab:artwork_rotowire} shows M$^2$EX outperforms CAESURA by 30\% on the ArtWork and by ca. 42\% on the RotoWire datasets in accuracy, with advantages in both single- and multi-modality queries. Efficiency-wise, M$^2$EX excels on ArtWork with fewer steps, lower latency, and reduced costs. On RotoWire, despite higher token usage and costs due to advanced text analysis, M$^2$EX maintains superior accuracy. Additionally, M$^2$EX supports re-planning and offers better explanations, features absent in CAESURA.\\
\noindent\textbf{Results on the EHRXQA Dataset}
In Table \ref{tab:ehr}, M$^2$EX outperforms NeuralSQL in overall accuracy (51.00\% vs. 33.00\% in 10-shot) on the EHRXQA dataset, especially in multiple-table queries (77.50\% vs. 47.50\%) and binary questions (74.00\% vs. 48.00\%). Additionally, M$^2$EX  provides plan generation (98\% coverage), explanations, and replanning—features that NeuralSQL lacks. Metrics like steps, tokens, and latency are excluded since NeuralSQL generates answers directly without intermediate steps, unlike M$^2$EX’s transparent workflow.
We exclude CAESURA from the EHRXQA experiments due to its inefficiency with EHRXQA’s complex schema. While CAESURA is intended to be a general-purpose multi-modal system, it processes the relational database through multiple steps, examining each table and relationship sequentially. This limitation introduces significant overhead when handling the complex data schema of the EHRXQA dataset (there are 18 tables) during its discovery phase. Consequently, reproducing CAESURA on EHRXQA questions fails to perform inferences at the early stages of the planning phase, ultimately terminating after exceeding the maximum number of allowed attempts.
\subsection{Error Analysis}
We evaluate system errors across three datasets: ArtWork, RotoWire, and EHRXQA, identifying key bottlenecks and component failures (see Table \ref{tab:error-headline} and detailed breakdown in Appendix \ref{err}, Fig. \ref{fig:error-analysis}). On the ArtWork dataset, CAESURA exhibits 20 errors out of 30 tasks, mainly due to faulty planning in sequential workflows and incorrect outputs from the image analysis module. Multi-modal tasks involving plot and data structure outputs are particularly error-prone, especially in parallel workflows where planning failures are common. By contrast, M$^2$EX achieves full planning success, with image interpretation errors being the only significant issue.
In the RotoWire dataset, CAESURA fails on 9 of 12 tasks due to text analysis failures and SQL generation flaws. M$^2$EX resolves all single-modal tasks but faces 4 errors in multi-modal tasks, again tied to text interpretation. These patterns highlight M$^2$EX’s robustness in planning and execution while exposing shared weaknesses in text and image understanding across systems.

For the EHRXQA dataset, we focus solely on M$^2$EX due to NeuralSQL’s lack of interpretable planning. Of 49 errors, 36 arise in categorical tasks, indicating a strong link between output type and model performance. Most failures originate from inaccurate image analysis by the M3AE model. These results emphasize the need for improved image understanding, especially for categorical reasoning, alongside stronger planning and SQL components. See Appendix \ref{err} for full error analysis.
\section{Conclusions}
% In this paper, we show that multi-agent collaboration using LLMs (GPT4o) is a promising approach for multi-modal data exploration in natural language. Our system, M$^2$EX, achieves superior accuracy and efficiency compared to state-of-the-art methods on datasets with tabular, text, and image data, leveraging smart re-planning and parallel execution. It also enhances transparency through detailed explanations, fostering user trust. Our work demonstrates an effective paradigm for integrating diverse data types, with strong performance in text-to-SQL tasks but room for improvement in image analysis and workflow optimization. 
% Future efforts should focus on exploring better data alignment, prompt engineering, planning optimization, scaling to larger datasets, and incorporating modalities like video and human-in-the-loop strategies. 
% Overall, M$^2$EX shows a significant advance in multi-modal data exploration, blending accuracy, efficiency, and user-centric design, with potential for further enhancement.
In this paper, we show that multi-agent collaboration via LLMs (GPT-4o) offers a powerful approach to multi-modal data exploration in natural language. Our system, M$^2$EX, outperforms prior methods across datasets with tabular, text, and image data by leveraging smart re-planning, parallel execution, and transparent, explainable workflows. It blends accuracy, efficiency, and user-centric design, marking a significant advance in multi-modal data exploration, with strong performance in text-to-SQL tasks and potential for further enhancement in image reasoning and workflow optimization.

Future work will focus on better data alignment, prompt design, planning efficiency, and scaling to larger datasets and new modalities such as video and human-in-the-loop interaction. 

\section*{Limitations}
Despite M$^2$EX's overall superior performance, several limitations remain. Most notably, the system's reliance on image analysis introduces a consistent source of error, particularly in tasks involving categorical outputs. The M3AE model often fails to capture subtle visual distinctions, which disproportionately affects the accuracy of multi-modal tasks. We did not explore alternative image processing approaches, as improving the visual pipeline was not the primary objective of this study. Instead, we adopted visual models commonly used in prior work to ensure a fair and consistent basis for comparison. Similarly, we restricted our language model experiments to GPT-4o to both showcase our proposed methods and maintain comparability with recent studies. 

Additionally, although M$^2$EX successfully generates plans for all tasks, its performance still hinges on accurate text interpretation. In the RotoWire dataset, for example, errors in multi-modal questions were largely driven by flawed text comprehension, revealing a vulnerability in the language understanding pipeline. 

Finally, the system exhibits a performance gap between binary and categorical tasks, suggesting that output type complexity influences success rates. These findings indicate that further improvements are needed in visual reasoning, nuanced language understanding, and output-type generalization.
\section*{Acknowledgments}
This project has received funding from the Hasler Foundation Project ID: 2024-05-21-76, DIZH Project-Call 2024.1, ID:9, and the DataGEMS, funded by European Union's Horizon Europe Research and Innovation programme, under grant agreement No 101188416.

\bibliographystyle{acl_natbib}
\bibliography{main}

\begin{thebibliography}{33}
\providecommand{\natexlab}[1]{#1}

\bibitem[{Bae et~al.(2024)Bae, Kyung, Ryu, Cho, Lee, Kweon, Oh, Ji, Chang, Kim et~al.}]{bae2024ehrxqa}
Seongsu Bae, Daeun Kyung, Jaehee Ryu, Eunbyeol Cho, Gyubok Lee, Sunjun Kweon, Jungwoo Oh, Lei Ji, Eric Chang, Tackeun Kim, and 1 others. 2024.
\newblock Ehrxqa: A multi-modal question answering dataset for electronic health records with chest x-ray images.
\newblock \emph{Advances in Neural Information Processing Systems}, 36.

\bibitem[{Chen et~al.(2022)Chen, Du, Hu, Liu, Li, Wan, and Chang}]{10.1007/978-3-031-16443-9_65}
Zhihong Chen, Yuhao Du, Jinpeng Hu, Yang Liu, Guanbin Li, Xiang Wan, and Tsung-Hui Chang. 2022.
\newblock \href {https://doi.org/10.1007/978-3-031-16443-9_65} {Multi-modal masked autoencoders for medical vision-and-language pretraining}.
\newblock In \emph{Medical Image Computing and Computer Assisted Intervention – MICCAI 2022: 25th International Conference, Singapore, September 18–22, 2022, Proceedings, Part V}, page 679–689, Berlin, Heidelberg. Springer-Verlag.

\bibitem[{Doe et~al.(2025)Doe, Smith, and Zhang}]{magma2025}
Jane Doe, John Smith, and Wei Zhang. 2025.
\newblock \href {https://example.com/magma-paper} {Magma: Multi-agent generalist for multimodal alignment}.
\newblock In \emph{Proceedings of the 2025 Conference on Empirical Methods in Natural Language Processing (EMNLP)}.
\newblock To appear.

\bibitem[{Dong et~al.(2024)Dong, Zhang, Zhou, Chen, Zha, and Huang}]{NEURIPS2024_d0aafec0}
Junnan Dong, Qinggang Zhang, Chuang Zhou, Hao Chen, Daochen Zha, and Xiao Huang. 2024.
\newblock \href {https://proceedings.neurips.cc/paper_files/paper/2024/file/d0aafec03d59db29a92fa683bd783374-Paper-Conference.pdf} {Cost-efficient knowledge-based question answering with large language models}.
\newblock In \emph{Advances in Neural Information Processing Systems}, volume~37, pages 115261--115281. Curran Associates, Inc.

\bibitem[{Du et~al.(2023)Du, Li, Tang, Zhao, and Wen}]{du-etal-2023-zero}
Yifan Du, Junyi Li, Tianyi Tang, Wayne~Xin Zhao, and Ji-Rong Wen. 2023.
\newblock \href {https://doi.org/10.18653/v1/2023.findings-acl.590} {Zero-shot visual question answering with language model feedback}.
\newblock In \emph{Findings of the Association for Computational Linguistics: ACL 2023}, pages 9268--9281, Toronto, Canada. Association for Computational Linguistics.

\bibitem[{Floratou et~al.(2024)Floratou, Psallidas, Zhao, Deep, Hagleither, Tan, Cahoon, Alotaibi, Henkel, Singla, van Grootel, Chow, Deng, Lin, Campos, Emani, Pandit, Shnayder, Wang, and Curino}]{Floratou2024}
Avrilia Floratou, Fotis Psallidas, Fuheng Zhao, Shaleen Deep, Gunther Hagleither, Wangda Tan, Joyce Cahoon, Rana Alotaibi, Jordan Henkel, Abhik Singla, Alex van Grootel, Brandon Chow, Kai Deng, Katherine Lin, Marcos Campos, Venkatesh Emani, Vivek Pandit, Victor Shnayder, Wenjing Wang, and Carlo Curino. 2024.
\newblock Nl2sql is a solved problem... not!
\newblock In \emph{CIDR}.

\bibitem[{F{\"u}rst et~al.(2024)F{\"u}rst, Kosten, Nooralahzadeh, Zhang, and Stockinger}]{furst2024evaluating}
Jonathan F{\"u}rst, Catherine Kosten, Farhad Nooralahzadeh, Yi~Zhang, and Kurt Stockinger. 2024.
\newblock Evaluating the data model robustness of text-to-sql systems based on real user queries.
\newblock \emph{arXiv preprint arXiv:2402.08349}.

\bibitem[{Gao et~al.(2025)Gao, Zhang, Li, Ma, Yuan, Fan, Wu, Jia, Zhu, and Li}]{gao2025multi}
Zhi Gao, Bofei Zhang, Pengxiang Li, Xiaojian Ma, Tao Yuan, Yue Fan, Yuwei Wu, Yunde Jia, Song-Chun Zhu, and Qing Li. 2025.
\newblock \href {https://openreview.net/forum?id=0bmGL4q7vJ} {Multi-modal agent tuning: Building a {VLM}-driven agent for efficient tool usage}.
\newblock In \emph{The Thirteenth International Conference on Learning Representations}.

\bibitem[{Jo and Trummer(2024)}]{jo2024}
Saehan Jo and Immanuel Trummer. 2024.
\newblock Thalamusdb: Approximate query processing on multi-modal data.
\newblock \emph{Proc. ACM Manag. Data}, 2(3).

\bibitem[{Kakkar et~al.(2023)Kakkar, Cao, Chunduri, Xu, Vyalla, Dintyala, Prabakaran, Bang, Sengupta, Ravichandran, Sivakumar, Rajoria, Raju, Aggarwal, Shah, Garg, Suman, Kalluraya, Mitra, Payani, Lu, Ramachandran, and Arulraj}]{DBLP:conf/deem/KakkarCCXVDPBS023}
Gaurav~Tarlok Kakkar, Jiashen Cao, Pramod Chunduri, Zhuangdi Xu, Suryatej~Reddy Vyalla, Prashanth Dintyala, Anirudh Prabakaran, Jaeho Bang, Aubhro Sengupta, Kaushik Ravichandran, Ishwarya Sivakumar, Aryan Rajoria, Ashmita Raju, Tushar Aggarwal, Abdullah Shah, Sanjana Garg, Shashank Suman, Myna~Prasanna Kalluraya, Subrata Mitra, and 4 others. 2023.
\newblock \href {https://doi.org/10.1145/3595360.3595858} {Eva: An end-to-end exploratory video analytics system}.
\newblock In \emph{Proceedings of the Seventh Workshop on Data Management for End-to-End Machine Learning}, DEEM '23, New York, NY, USA. Association for Computing Machinery.

\bibitem[{Kang et~al.(2019)Kang, Bailis, and Zaharia}]{DBLP:journals/pvldb/KangBZ19}
Daniel Kang, Peter Bailis, and Matei Zaharia. 2019.
\newblock Blazeit: Optimizing declarative aggregation and limit queries for neural network-based video analytics.
\newblock \emph{Proc. {VLDB} Endow.}, 13(4):533--546.

\bibitem[{Kang et~al.(2022)Kang, Romero, Bailis, Kozyrakis, and Zaharia}]{DBLP:conf/cidr/KangRBKZ22}
Daniel Kang, Francisco Romero, Peter~D. Bailis, Christos Kozyrakis, and Matei Zaharia. 2022.
\newblock {VIVA:} an end-to-end system for interactive video analytics.
\newblock In \emph{CIDR}.

\bibitem[{Kapoor et~al.(2024)Kapoor, Stroebl, Siegel, Nadgir, and Narayanan}]{kapoor2024ai}
Sayash Kapoor, Benedikt Stroebl, Zachary~S Siegel, Nitya Nadgir, and Arvind Narayanan. 2024.
\newblock Ai agents that matter.
\newblock \emph{arXiv preprint arXiv:2407.01502}.

\bibitem[{Kim et~al.(2023)Kim, Moon, Tabrizi, Lee, Mahoney, Keutzer, and Gholami}]{kim2023llm}
Sehoon Kim, Suhong Moon, Ryan Tabrizi, Nicholas Lee, Michael~W Mahoney, Kurt Keutzer, and Amir Gholami. 2023.
\newblock An llm compiler for parallel function calling.
\newblock \emph{arXiv preprint arXiv:2312.04511}.

\bibitem[{Ko et~al.(2023)Ko, Lee, Kang, Roh, and Kim}]{ko-etal-2023-large}
Dohwan Ko, Ji~Lee, Woo-Young Kang, Byungseok Roh, and Hyunwoo Kim. 2023.
\newblock \href {https://doi.org/10.18653/v1/2023.emnlp-main.261} {Large language models are temporal and causal reasoners for video question answering}.
\newblock In \emph{Proceedings of the 2023 Conference on Empirical Methods in Natural Language Processing}, pages 4300--4316, Singapore. Association for Computational Linguistics.

\bibitem[{Lewis et~al.(2020)Lewis, Liu, Goyal, Ghazvininejad, Mohamed, Levy, Stoyanov, and Zettlemoyer}]{lewis2020bart}
Mike Lewis, Yinhan Liu, Naman Goyal, Marjan Ghazvininejad, Abdelrahman Mohamed, Omer Levy, Veselin Stoyanov, and Luke Zettlemoyer. 2020.
\newblock Bart: Denoising sequence-to-sequence pre-training for natural language generation, translation, and comprehension.
\newblock In \emph{Proceedings of the 58th Annual Meeting of the Association for Computational Linguistics}, page 7871. Association for Computational Linguistics.

\bibitem[{Li et~al.(2024{\natexlab{a}})Li, Zhang, Zhang, Guo, Zhang, Li, Zhang, Liu, and Li}]{li2024llavanext-strong}
Bo~Li, Kaichen Zhang, Hao Zhang, Dong Guo, Renrui Zhang, Feng Li, Yuanhan Zhang, Ziwei Liu, and Chunyuan Li. 2024{\natexlab{a}}.
\newblock \href {https://llava-vl.github.io/blog/2024-05-10-llava-next-stronger-llms/} {Llava-next: Stronger llms supercharge multimodal capabilities in the wild}.

\bibitem[{Li et~al.(2024{\natexlab{b}})Li, Hui, Qu, Yang, Li, Li, Wang, Qin, Geng, Huo et~al.}]{li2024can}
Jinyang Li, Binyuan Hui, Ge~Qu, Jiaxi Yang, Binhua Li, Bowen Li, Bailin Wang, Bowen Qin, Ruiying Geng, Nan Huo, and 1 others. 2024{\natexlab{b}}.
\newblock Can llm already serve as a database interface? a big bench for large-scale database grounded text-to-sqls.
\newblock \emph{NeurIPS}.

\bibitem[{Li et~al.(2023{\natexlab{a}})Li, Li, Savarese, and Hoi}]{10.5555/3618408.3619222}
Junnan Li, Dongxu Li, Silvio Savarese, and Steven Hoi. 2023{\natexlab{a}}.
\newblock Blip-2: bootstrapping language-image pre-training with frozen image encoders and large language models.
\newblock In \emph{Proceedings of the 40th International Conference on Machine Learning}, ICML'23. JMLR.org.

\bibitem[{Li et~al.(2023{\natexlab{b}})Li, Li, Savarese, and Hoi}]{li2023blip}
Junnan Li, Dongxu Li, Silvio Savarese, and Steven Hoi. 2023{\natexlab{b}}.
\newblock Blip-2: Bootstrapping language-image pre-training with frozen image encoders and large language models.
\newblock In \emph{International conference on machine learning}, pages 19730--19742. PMLR.

\bibitem[{Liu et~al.(2024{\natexlab{a}})Liu, Russo, Cafarella, Cao, Baille~Chen, Chen, Franklin, Kraska, Madden, and Vitagliano}]{liu2024declarative}
Chunwei Liu, Matthew Russo, Michael Cafarella, Lei Cao, Peter Baille~Chen, Zui Chen, Michael Franklin, Tim Kraska, Samuel Madden, and Gerardo Vitagliano. 2024{\natexlab{a}}.
\newblock A declarative system for optimizing ai workloads.
\newblock \emph{arXiv e-prints}, pages arXiv--2405.

\bibitem[{Liu et~al.(2024{\natexlab{b}})Liu, Hill, Du, Wang, and Tong}]{liu-etal-2024-conversational}
Lihui Liu, Blaine Hill, Boxin Du, Fei Wang, and Hanghang Tong. 2024{\natexlab{b}}.
\newblock \href {https://doi.org/10.18653/v1/2024.findings-acl.48} {Conversational question answering with language models generated reformulations over knowledge graph}.
\newblock In \emph{Findings of the Association for Computational Linguistics: ACL 2024}, pages 839--850, Bangkok, Thailand. Association for Computational Linguistics.

\bibitem[{Nooralahzadeh et~al.(2024)Nooralahzadeh, Zhang, Smith, Maennel, Matthey-Doret, de~Fondville, and Stockinger}]{noor2024}
Farhad Nooralahzadeh, Yi~Zhang, Ellery Smith, Sabine Maennel, Cyril Matthey-Doret, Raphaël de~Fondville, and Kurt Stockinger. 2024.
\newblock {StatBot.Swiss: Bilingual Open Data Exploration in Natural Language}.
\newblock In \emph{Findings of ACL}.

\bibitem[{Pourreza and Rafiei(2024)}]{pourreza2024din}
Mohammadreza Pourreza and Davood Rafiei. 2024.
\newblock Din-sql: Decomposed in-context learning of text-to-sql with self-correction.
\newblock \emph{NeurIPS}.

\bibitem[{Schick et~al.(2023)Schick, Dwivedi-Yu, Dessi, Raileanu, Lomeli, Hambro, Zettlemoyer, Cancedda, and Scialom}]{schick_toolformer_2023}
Timo Schick, Jane Dwivedi-Yu, Roberto Dessi, Roberta Raileanu, Maria Lomeli, Eric Hambro, Luke Zettlemoyer, Nicola Cancedda, and Thomas Scialom. 2023.
\newblock \href {https://proceedings.neurips.cc/paper_files/paper/2023/hash/d842425e4bf79ba039352da0f658a906-Abstract-Conference.html} {Toolformer: {Language} {Models} {Can} {Teach} {Themselves} to {Use} {Tools}}.
\newblock \emph{Advances in Neural Information Processing Systems}, 36:68539--68551.

\bibitem[{Sivasubramaniam et~al.(2024)Sivasubramaniam, Osei-Akoto, Zhang, Stockinger, and Fuerst}]{sivasubramaniamsm3}
Sithursan Sivasubramaniam, Cedric Osei-Akoto, Yi~Zhang, Kurt Stockinger, and Jonathan Fuerst. 2024.
\newblock \href {https://openreview.net/forum?id=Pm0UzCehgB} {{SM}3-text-to-query: Synthetic multi-model medical text-to-query benchmark}.
\newblock In \emph{The Thirty-eight Conference on Neural Information Processing Systems Datasets and Benchmarks Track}.

\bibitem[{Urban and Binnig(2024)}]{urbanB24}
Matthias Urban and Carsten Binnig. 2024.
\newblock {CAESURA:} language models as multi-modal query planners.
\newblock In \emph{CIDR}.

\bibitem[{Yang et~al.(2023)Yang, Li, Wang, Lin, Azarnasab, Ahmed, Liu, Liu, Zeng, and Wang}]{yang2023mmreact}
Zhengyuan Yang, Linjie Li, Jianfeng Wang, Kevin Lin, Ehsan Azarnasab, Faisal Ahmed, Zicheng Liu, Ce~Liu, Michael Zeng, and Lijuan Wang. 2023.
\newblock Mm-react: Prompting chatgpt for multimodal reasoning and action.

\bibitem[{Yao et~al.(2023)Yao, Zhao, Yu, Du, Shafran, Narasimhan, and Cao}]{yao2023react}
Shunyu Yao, Jeffrey Zhao, Dian Yu, Nan Du, Izhak Shafran, Karthik~R Narasimhan, and Yuan Cao. 2023.
\newblock \href {https://openreview.net/forum?id=WE_vluYUL-X} {React: Synergizing reasoning and acting in language models}.
\newblock In \emph{The Eleventh International Conference on Learning Representations}.

\bibitem[{Yu et~al.(2018)Yu, Zhang, Yang, Yasunaga, Wang, Li, Ma, Li, Yao, Roman et~al.}]{yu2018spider}
Tao Yu, Rui Zhang, Kai Yang, Michihiro Yasunaga, Dongxu Wang, Zifan Li, James Ma, Irene Li, Qingning Yao, Shanelle Roman, and 1 others. 2018.
\newblock Spider: A large-scale human-labeled dataset for complex and cross-domain semantic parsing and text-to-sql task.
\newblock In \emph{EMNLP}.

\bibitem[{Zhang et~al.(2023)Zhang, Daum, He, Haynes, Krishna, and Balazinska}]{zhang2023equi}
Enhao Zhang, Maureen Daum, Dong He, Brandon Haynes, Ranjay Krishna, and Magdalena Balazinska. 2023.
\newblock Equi-vocal: Synthesizing queries for compositional video events from limited user interactions.
\newblock \emph{Proceedings of the VLDB Endowment}, 16(11):2714--2727.

\bibitem[{Zhang et~al.(2024{\natexlab{a}})Zhang, Huang, Jin, and Lu}]{zhang2024vision}
Jingyi Zhang, Jiaxing Huang, Sheng Jin, and Shijian Lu. 2024{\natexlab{a}}.
\newblock Vision-language models for vision tasks: A survey.
\newblock \emph{IEEE Transactions on Pattern Analysis and Machine Intelligence}.

\bibitem[{Zhang et~al.(2024{\natexlab{b}})Zhang, Deriu, Katsogiannis-Meimarakis, Kosten, Koutrika, and Stockinger}]{zhang2023sciencebenchmark}
Yi~Zhang, Jan Deriu, George Katsogiannis-Meimarakis, Catherine Kosten, Georgia Koutrika, and Kurt Stockinger. 2024{\natexlab{b}}.
\newblock Sciencebenchmark: A complex real-world benchmark for evaluating natural language to sql systems.
\newblock \emph{Proceedings of the VLDB Endowment}, 17(4):685--698.

\end{thebibliography}

\appendix
\onecolumn
\section{M$^2$EX on ArtWork}\label{XMODE-artwork}
\vspace{-10pt}
\begin{figure}[h!]
    \centering
\includegraphics[width=.9\textwidth]{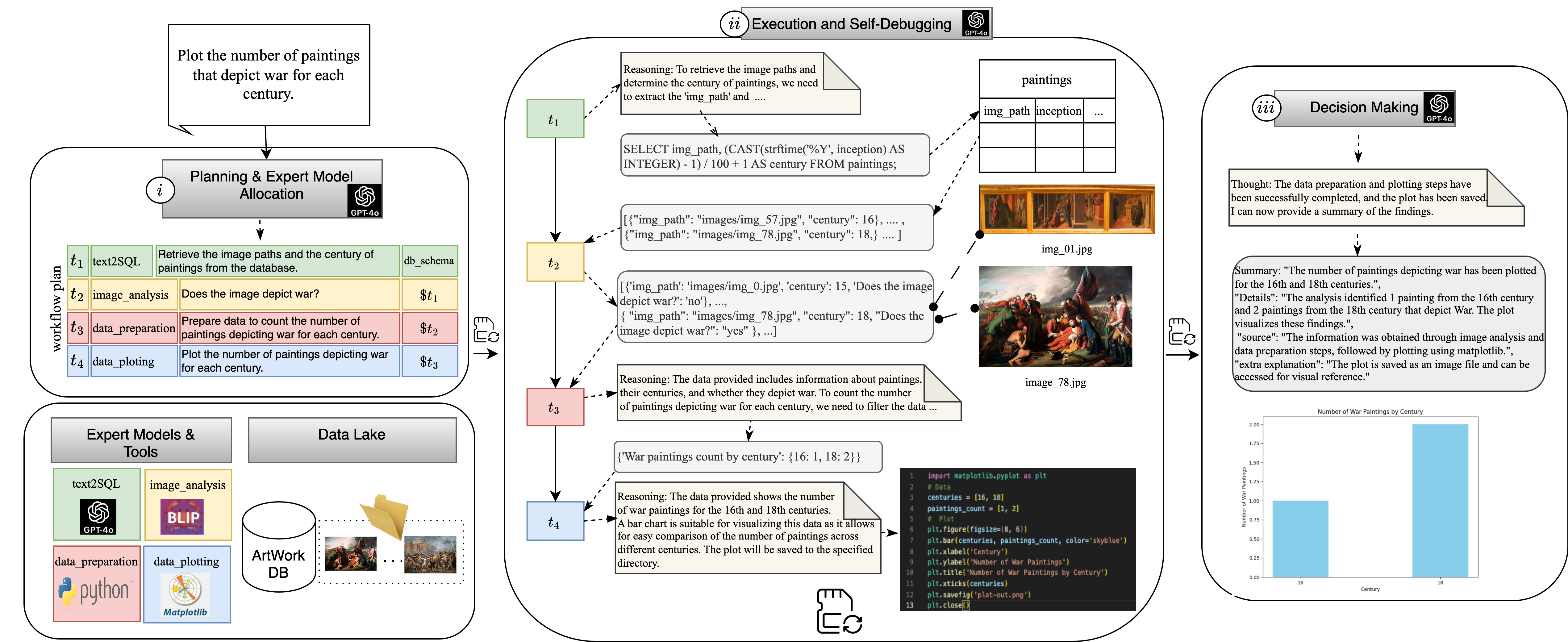}
\vspace{-10pt}

\caption{M$^2$EX framework on ArtWork ~\citep{urbanB24} with an example of processing a multi-modal query. The query is automatically decomposed into various components such as text2SQL, and image analysis which can be inspected by the user for explainability.}
\label{fig:XMODE-artwork}
\end{figure}

\section{Prompts}\label{prompts}
\vspace{-10pt}
\begin{tcolorbox}[colback=gray!10, colframe=black, width=1\textwidth, 
                  sharp corners, boxrule=1pt, title= Planner Prompt / Replanning Prompt]
\tiny
\ttfamily % Use monospaced font
[SYSTEM]: Given a user question and a database schema, analyze the question to identify and break it down into relevant sub-questions. 

Determine which tools (e.g., \textcolor{blue}{\{tool\_names\}}) are appropriate for answering each sub-question based on the available database information and tools.
    
Decompose the user question into sub-questions that capture all elements of the question’s intent. This includes identifying the main objective, relevant sub-questions, necessary background information, assumptions, and any secondary requirements. 
    
Ensure that no part of the original question’s intent is omitted, and create a list of individual steps to answer the question fully and accurately using tools. 
    
You may need to use one tool multiple times to answer the original question.

First, you should begin by thoroughly analyzing the user's main question. It’s important to understand the key components and objectives within the query.

Next, you must review the provided database schema. This involves examining the tables, fields, and relationships within the database to identify which parts of the schema are relevant to the user’s question and contribute to a set of sub-questions.

For each sub-question, provide all the required information that may required in other tasks. In order to find this information look at the user question and the database information.

Each sub-question or step should focus exclusively on a single task.    

Each sub-question should be a textual question. Don't generate a code as a sub-question.

Create a plan to solve it with the utmost parallelizability. 

Each plan should comprise an action from the following 
\textcolor{blue}{\{num\_tools\} }types:

\textcolor{blue}{\{tool\_descriptions\}}

\textcolor{blue}{\{num\_tools\}}. join(): Collects and combines results from prior actions.

 - An LLM agent is called upon invoking join() to either finalize the user query or wait until the plans are executed.
 
 - join should always be the last action in the plan, and will be called in two scenarios:
 
   (a) if the answer can be determined by gathering the outputs from tasks to generate the final response.
   
   (b) if the answer cannot be determined in the planning phase before you execute the plans. Guidelines:
   
 - Each action described above contains input/output types and descriptions.
 
- You must strictly adhere to the input and output types for each action.
    
- The action descriptions contain the guidelines. You MUST strictly follow those guidelines when you use the actions.
    
- Each action in the plan should strictly be one of the above types. Follow the Python conventions for each action.

- Each action MUST have a unique ID, which is strictly increasing.

- Inputs for actions can either be constants or outputs from preceding actions. In the latter case, use the format  \$id to denote the ID of the previous action whose output will be the input.

- If there is an input from preceding actions, always point its id as `\$id` in the context of the action

- Always call join as the last action in the plan. Say '<END\_OF\_PLAN>' after you call join.

 - Ensure the plan maximizes parallelizability.
 
 - Only use the provided action types. If a query cannot be addressed using these, invoke the join action for the next steps.
 
 - Never introduce new actions other than the ones provided.

\textcolor{blue}{\{list of usecase-specific business rules}\}

[USER]:\textcolor{blue}{\{state\}}
% [{HumanMessage\{\{question\}, \{db schema\}]\}\}}}

[SYSTEM]: Remember, ONLY respond with the task list in the correct format! E.g.: idx. tool(arg\_name=args),

\end{tcolorbox}

\begin{tcolorbox}[colback=gray!10, colframe=black, width=1\textwidth, 
                  sharp corners, boxrule=1pt, title= Prompt for Decision Making ]
\tiny
\ttfamily % Use monospaced font
[SYSTEM]: Solve a question answering task. Here are some guidelines:

- In the Assistant Scratchpad, you will be given results of a plan you have executed to answer the user's question.

    - Thought needs to reason about the question based on the Observations in 1-2 sentences.
    
    - Ignore irrelevant action results.
    
    - If the required information is present, give a concise but complete and helpful answer to the user's question.
    - If you are unable to give a satisfactory finishing answer, replan to get the required information. Respond in the following format:
    
    Thought: <reason about the task results and whether you have sufficient information to answer the question>
    
    Action: <action to take>
    
    - If an error occurs during previous actions, replan and take corrective measures to obtain the required information.
    
    - Ensure that you consider errors in all the previous steps, and try to replan accordingly.
    
    - Ensure the final answer is provided in a structured format as JSON as follows:
    
        \{\{'Summary': <concise summary of the answer>,
        
         'details': <detailed explanation and supporting information>,
         
         'source': <source of the information or how it was obtained>,
         
         'inference':<your final inference as YES, No, or list of requested information without any extra information which you can take from the `labels` as given below>,
         'extra explanation':<put here the extra information that you don't provide in inference >,
         
         \}\}
         
         In the `inference` do not provide additional explanation or description. Put them in `extra explanation`.

    Available actions:
    
    (1) Finish (the final answer to return to the user): returns the answer and finishes the task.
    
    (2) Replan(the reasoning and other information that will help you plan again. Can be a line of any length): instructs why we must replan.

    [USER]: \textcolor{blue}{\{state\}}
    % [HumanMessage:\{...\}, [FunctionMessage: \{...\}, ...]]\}}
        
    [SYSTEM]: Using the above previous actions, decide whether to replan or finish. 
    
        If all the required information is present, you may finish. Consider replanning for data\_preparation task if you want to structure the response in a proper way.
        
        If you have made many attempts to find the information without success, admit so and respond with whatever information you have gathered so the user can work well with you.
        
        Do not generate a response based on the sample data (assumption). If you failed after multiple attempts, you can finish and explain the reason.
\end{tcolorbox}

\begin{tcolorbox}[colback=gray!10, colframe=black, width=1\textwidth, 
                  sharp corners, boxrule=1pt, title= Prompt for text2SQL ]
\tiny
\ttfamily % Use monospaced font
[SYSTEM]:  You are a database expert. Generate a SQL query given the following user question, database information and other context that you receive.
You should analyse the question, context and database schema and come up with the executable sqlite3 query. 

Provide all the required information in the SQL code to answer the original user question that may required in other tasks utilizing the relevant database schema.

Ensure you include all necessary information, including columns used for filtering, especially when the task involves plotting or data exploration.

This must be taken into account when performing any time-based data queries or analyses.

Translate a text question into a SQL query that can be executed on the SQLite database.

You should stick to the available schema including tables and columns in the database and should not bring any new tables or columns.

 [USER]: \textcolor{blue}{\{text2SQL task description\}, \{db schema\}}
\end{tcolorbox}

% \begin{tcolorbox}[colback=gray!10, colframe=black, width=1\textwidth, 
%                   sharp corners, boxrule=1pt, title= Prompt for image\_analysis ]
% \tiny
% \ttfamily % Use monospaced font
% [SYSTEM]:  
% \end{tcolorbox}

\begin{tcolorbox}[colback=gray!10, colframe=black, width=1\textwidth, 
                  sharp corners, boxrule=1pt, title= Prompt for text\_analysis ]
\tiny
\ttfamily % Use monospaced font
[SYSTEM]:  You are a text analysis assistant. 
Analyze the provided question and report to answer the question. 

Only answer the question and don't provide extra information in your answer. 

In your answer, be concrete and use None if you can't find the answer in the report.

The output should be in the format: \{\{'reasoning': '...', 'answer': '...'\}\}

[USER]: \textcolor{blue}{\{text analysis task description\}, \{text\}}
\end{tcolorbox}

\begin{tcolorbox}[colback=gray!10, colframe=black, width=1\textwidth, 
                  sharp corners, boxrule=1pt, title= Prompt for data\_preparation ]
\tiny
\ttfamily % Use monospaced font
[SYSTEM]:  You are a data preparation and processing assistant. Create a proper structure for the provided data from the previous steps to answer the request.

- If the required information has not found in the provided data, ask for replanning and ask from previous tools to include the missing information.

- You should include all the input data in the code, and prevent of ignoring them by  `\# ... (rest of the data)`.

- You should provide a name or caption for each value in the final output considering the question and the input context."

- Don't create any sample data in order to answer to the user question.

- You should print the final data structure.

- You should save the final data structure at the specified path with a proper filename.

- You should output the final data structure as a final output.

 [USER]: \textcolor{blue}{\{data preparation task description\}, \{result from previous task\}}
\end{tcolorbox}

\begin{tcolorbox}[colback=gray!10, colframe=black, width=1\textwidth, 
                  sharp corners, boxrule=1pt, title= Prompt for data\_plotting]
\tiny
\ttfamily % Use monospaced font
[SYSTEM]:  You are a data plotting assistant. Plot the provided data from the previous steps to answer the question.

- Analyze the user's request and input data to determine the most suitable type of visualization/plot that also can be understood by the simple user.

- If the required information has not been found in the provided data, ask for replanning and ask from previous tools to include the missing information.

- Don't create any sample data in order to answer to the user question.

- You should save the generated plot at the specified path with the proper filename and .png extension.

 [USER]: \textcolor{blue}{\{data plotting task description\}, \{data\}}
\end{tcolorbox}
\vspace{2cm}

\section{Tools, Models, and Prompts by Subtask }\label{sec:appendix-tools}
\begin{table}[h]
\centering

\small
\setlength{\tabcolsep}{6pt}

\begin{tabular}{@{}p{5cm} p{5cm} p{4cm}@{}}
\toprule
\textbf{Task} & \textbf{Tool / Model} & \textbf{Prompt Type} \\
\midrule
Text-to-SQL translation          & GPT-4o                & text2SQL prompt \\
Text analysis     & GPT-4o                & text\_analysis prompt \\
ArtWork VQA         & BLIP-2                & no prompt \\
Medical image (EHRXQA)  VQA      & M3AE                  &  no prompt\\
Data preparation          &  GPT-4o and Python (Pandas)       & data\_preparation pormpt and Code via LLM output \\
Plot generation                  & GPT-4o and Matplotlib + Pandas   & data\_plotting prompt and Chart Code via LLM output \\
DAG construction (planning/replannig)      & GPT-4o (Planner loop) & planner Prompt / replanning  prompt \\
Decision Making & GPT-4o&  decision making prompt \\
%Tool metadata extraction         & Meta-prompt script    & Auto-generated from docstrings \\
\bottomrule
\end{tabular}
\caption{Subtasks, their associated tools/models, and prompt styles used in M$^2$EX. Most tool invocations are zero-shot or template-based.}
\label{tab:tool-table}
\end{table}

\section{Optimizations in M$^2$EX Explained with Examples}\label{opts}

\begin{figure*}[h]
    \centering
    %\subfloat[]{\includegraphics[width=0.9\textwidth]{figures/XMODE-smart-replaning.png}} %\hspace{0.01cm}
     \includegraphics[width=0.7\textwidth]{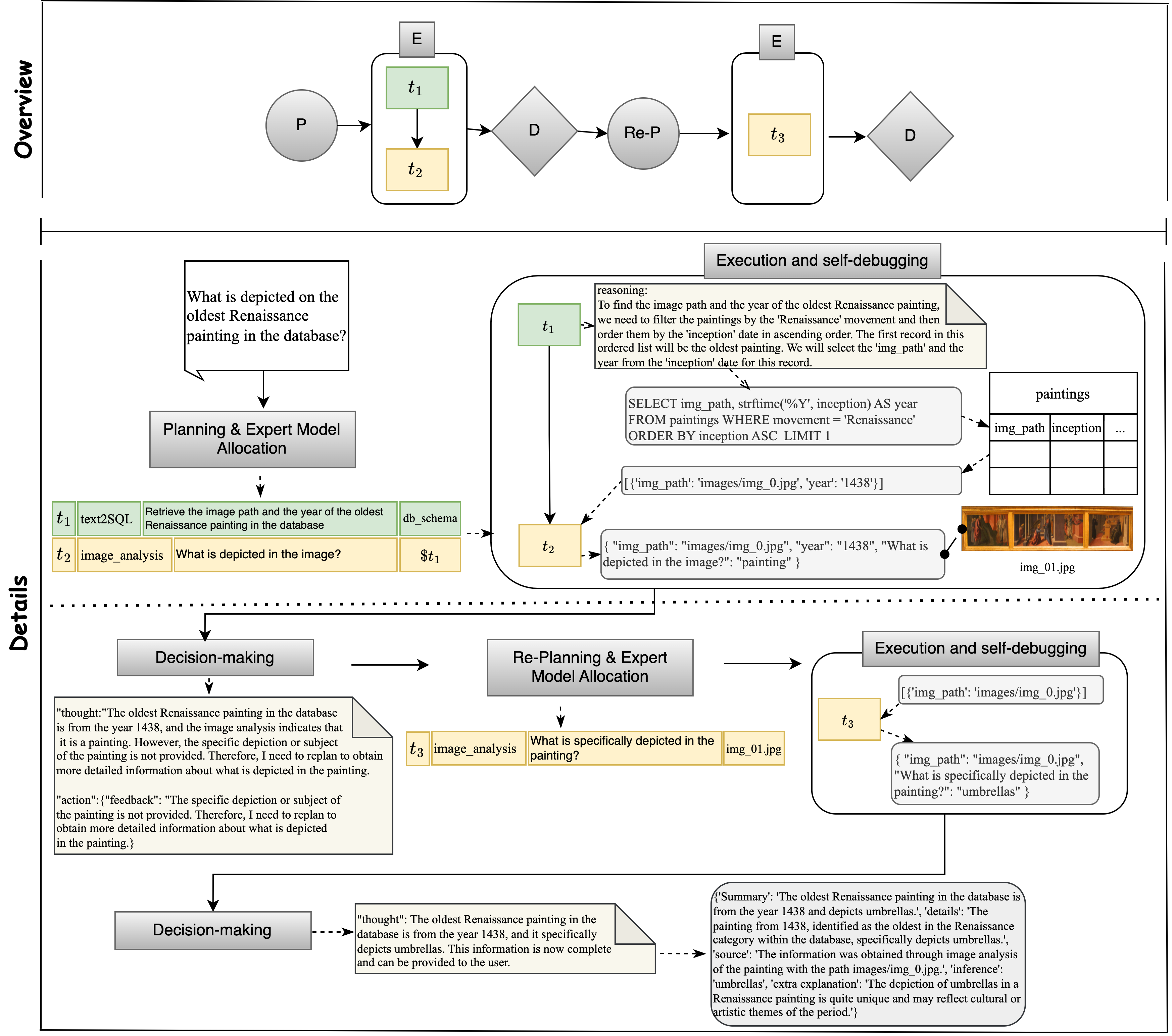} %\hspace{0.01cm}
%    \subfloat[Parallel planning]{\includegraphics[width=.5\textwidth]{figures/XMODE-conurrent.png}}
    \caption{Optimization of M$^2$EX: Smart replanning.}
    \label{fig:XMODE-planning}
\end{figure*}

To better demonstrate advantages of M$^2$EX, we provide several examples (see Figures \ref{fig:XMODE-artwork}  and \ref{fig:XMODE-planning}) across three key aspects: \textit{explanations, smart replanning, and parallel planning}. The following examples provide a detailed illustration of these three aspects.

\begin{mdframed}[hidealllines=true,backgroundcolor=cyan!20,innerleftmargin=3pt,innerrightmargin=3pt,leftmargin=-1pt,rightmargin=-1pt]

Example 1: \textit{Plot the number of paintings that depict war for each century} (see Figure \ref{fig:XMODE-artwork}).
\end{mdframed}

Through a series of well-planned and systematically executed steps, the model demonstrates not only how it processes the query but also how it provides transparency and reasoning at every stage, ensuring the user understands the process and results. The figure depicts a workflow that involves (1) Planning \& Expert Model Allocation, (2) Execution \& Self-Debugging, and (3) Decision Making. Here’s a breakdown of each step:\\
\textit{1) Planning \& Expert Model Allocation}: The process begins with the query being broken down into a sequence of subtasks: 
Task 1: Retrieve painting metadata, including their years and associated centuries, from the database.
Task 2: Analyze the images to determine whether they depict war.
Task 3: Prepare the data by counting the number of war-related paintings per century.
Task 4: Visualize these counts in a bar chart. 

Each task is allocated to specialized tools or models, such as text2SQL to translate the natural language question to SQL and database retrieval, image analysis tools for visual interpretation, coding tools to structure the data, and visualization libraries like matplotlib. This stage establishes a clear plan, showing how the overall query will be tackled in logical steps. \\
\textit{2) Execution \& Self-Debugging}:
The model begins executing the tasks, providing explanations and outputs at every stage to ensure clarity.
Task 1 - Retrieving Data:
The model constructs a SQL query to retrieve the required information from the database. It explains its reasoning: to determine the century of each painting, it converts the inception year into century values. The result is a list of paintings, each associated with its image path and century. 
Task 2 - Image Analysis:
With the retrieved data, the model analyzes each painting to determine if it depicts war. It applies image analysis tools to interpret the visual content of the paintings. The reasoning here is clear—war-related imagery, such as battles or soldiers, must be identified to answer the query. The output is a dataset indicating whether each painting depicts war.
Task 3 - Data Preparation:
The model filters and aggregates the data, counting the number of paintings depicting war for each century. It explains that grouping the paintings by century allows for easy comparison of trends across time periods. The result is a concise summary: \texttt{1 painting from the 16th century and 2 from the 18th century are identified as depicting war.}
Task 4 - Data Visualization:
Finally, the model prepares a bar chart to visualize the results. It explains its reasoning for choosing this visualization: bar charts effectively compare counts across categories, in this case, centuries. A Python script is provided, showing how the chart was generated, and the output is saved as an image for user reference.\\
\textit{3) Decision Making}: When the tasks are completed, the model reflects on its work and provides a final output based on its thought as \texttt{Summary:"The number of paintings depicting war has been plotted for the 16th and 18th centuries.", 
"Details": "The analysis identified 1 painting from the 16th century and 2 paintings from the 18th century that depict war. The plot visualizes these findings. [..]}". Throughout the workflow, the model demonstrates a commitment to transparency.

At every stage, M$^2$EX provides reasoning to justify its actions, from choosing SQL for retrieval to selecting a bar chart for visualization.
Intermediate outputs, like the dataset of war paintings and the Python plotting code, are made visible, ensuring the user can trace the steps taken.
The decision making phase wraps up the process by summarizing findings, clarifying the approach, and sharing the final visual result.
This shows that M$^2$EX not only answers the query effectively but also ensures its steps are understandable, logical, and well-documented, building trust in its analysis.
\begin{figure*}[h]
    \centering
%    \subfloat[Smart replanning]{\includegraphics[width=0.5\textwidth]{figures/XMODE-smart-replaning.png}} %\hspace{0.01cm}
    %\subfloat[]{\includegraphics[width=.9\textwidth]{figures/XMODE-conurrent.png}}
    \includegraphics[width=.695\textwidth]{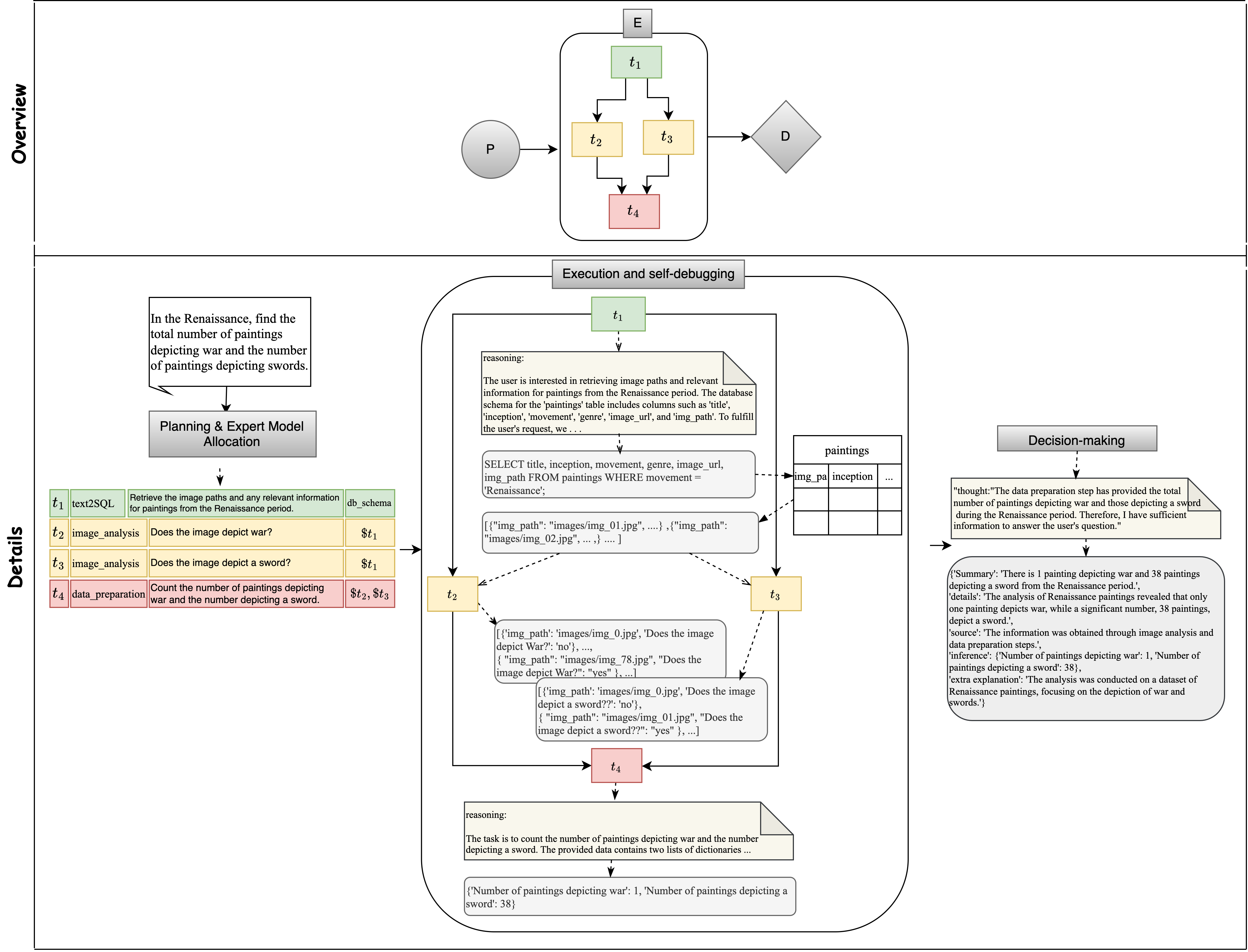}
    \caption{Optimization of M$^2$EX: Parallel planning.}
    \label{fig:XMODE-planning2}
\end{figure*}

\begin{mdframed}[hidealllines=true,backgroundcolor=cyan!20,innerleftmargin=2pt,innerrightmargin=3pt,leftmargin=-1pt,rightmargin=-1pt]
Example 2 - Smart Replanning: \textit{What is depicted on the oldest Renaissance painting in the database?} (see Figure \ref{fig:XMODE-planning}).
\end{mdframed}
Contrary to the previous example, M$^2$EX here involves \textit{smart replanning} - a major optimization technique of M$^2$EX. The main idea is to dynamically adapt the planning in case some tasks of the workflow fail or do not produce any results. Here’s a breakdown of each step:\\
\textit{1) Planning \& Expert Model Allocation}: M$^2$EX outputs the initial workflow plan that has 2 tasks. The first task involves retrieving the image path and the year of the oldest Renaissance painting in the database using a "text2SQL" expert model. It also involves an "image\_analysis" expert model in the second task, which aims to determine what is depicted in the image. \\
\textit{2) Execution and Self-Debugging}: M$^2$EX takes the information about the planned workflow as well as task dependencies and puts it into action. In Task 1, it comes with a reasoning statement to generate the SQL query as: \texttt{SELECT img\_path, strftime('\%Y', inception) AS year  FROM paintings WHERE movement = 'Renaissance' ORDER BY inception ASC  LIMIT 1}. Then it executes the query over the Artwork database and retrieves the specific image path and year for the oldest Renaissance painting as \texttt{[{'img\_path': 'images/img\_0.jpg', 'year': '1438'}]}. This allows the model to access the actual painting data in the subsequent  task. 

In Task 2, M$^2$EX utilizes the "image\_analysis" expert model (i.e. visual question answering based on BLIP) to examine the contents of  \texttt{img\_0.jpg} to answer the question: \textit{What is depicted in the image?}  The output of this task is transferred as a final result to the decision making component. At this point, the model's "thought" process in this component becomes evident. It reasons that while it knows that \texttt{img\_0.jpg} is a painting, the details about what is depicted in the painting have not been provided. Therefore, the model decides to not provide a final answer to the user and does replanning. 

The replanning capability is a crucial aspect of the M$^2$EX's approach. Rather than blindly accepting the final answer which does not produce a satisfiable or correct result, the model recognizes the need to replan and calls the "image\_analysis" module again. Since the model already knows which image in the database contains the oldest Renaissance painting, it smartly plans the "image\_analysis" task as Task 3, by reformulating the question as \textit{What is specifically depicted in the painting?} M$^2$EX then executes the task, and receives the more concrete answer "umbrellas". 

Moving forward, the decision making component confirms the details about the painting. Here, it verifies that the information it has gathered so far aligns with the natural language question and makes sense as a comprehensive understanding of the oldest Renaissance painting. The key aspect is the model's ability to replan effectively and to strategically leverage the available information to avoid repeating tasks.
\begin{mdframed}[hidealllines=true,backgroundcolor=cyan!20,innerleftmargin=3pt,innerrightmargin=3pt,leftmargin=-1pt,rightmargin=-1pt]
Example 3 - Parallel Planning: \textit{In the Renaissance, find the total number of paintings depicting war and the number of paintings depicting swords} (see Figure \ref{fig:XMODE-planning2}).
\end{mdframed}
The figure illustrates how M$^2$EX processes a complex query about Renaissance paintings, focusing on identifying how many paintings depict war and how many depict swords. The pipeline is structured to combine \textit{parallel task execution with step-by-step explanations}, ensuring clarity and efficiency throughout the process.

The process begins in the Planning \& Expert Model Allocation, where the model breaks down the user's query into distinct subtasks. These subtasks are assigned to specialized modules: 
Task 1 "text2SQL": This task retrieves image paths and relevant metadata for Renaissance paintings from a database using a SQL query.
Task 2 "image\_analysis": This task examines whether each painting depicts war.
Task 3 "image\_analysis": Simultaneously, another module analyzes whether each painting depicts a sword.
Task 4 "data\_preparation": This task consolidates the results from Task 2 and Task 3 to count and summarize the paintings.

The execution phase begins with Task 1, where the model generates and runs a SQL query. The reasoning provided for this step explains how the schema is understood and how the query ensures that only Renaissance paintings are retrieved. The output of Task 1 includes image paths and metadata, which are then sent to the next stage.

At this point, the model showcases its parallel planning capability. Tasks 2 and 3 are performed concurrently:
For Task 2, the system uses image analysis to determine if each painting depicts war.
For Task 3, a similar image analysis process identifies paintings that depict swords. Running these tasks in parallel significantly speeds up the workflow, as they operate independently of each other.
Once the image analysis tasks are complete, the model transitions to Task 4, where it aggregates the results. The reasoning here details how the system compiles two lists - one for paintings depicting war and one for those depicting swords. Afterwards, M$^2$EX counts the entries in each list. The final results are prepared for the decision making module.

In the decision making phase, the model reflects on its findings. It confirms that sufficient data was processed to answer the query and provides a summary: \texttt{"There is 1 painting depicting war and 38 paintings depicting swords.}"

M$^2$EX offers details, explaining how the analysis was conducted and highlighting the disparity between the two categories of paintings.
The system further provides an explanation of its methodology, emphasizing how it worked systematically to answer the query.
This demonstrates M$^2$EX's ability to manage tasks efficiently through parallel execution and to ensure transparency through reasoned explanations at every step. By combining these capabilities, the system provides a clear, accurate, and well-supported response to the user’s query.

Note that we did not compare M$^2$EX with NeuralSQL on ArtWork dataset, as such a comparison would be unfair due to NeuralSQL's inability to support plotting.

\section{Error Analysis}\label{err}
\vspace{-10pt}

\begin{figure}[h]
    \centering
    \subfloat[CAESURA (ArtWork)]{\includegraphics[trim={25pt 25pt 25pt 32pt},clip, width=0.42\linewidth]{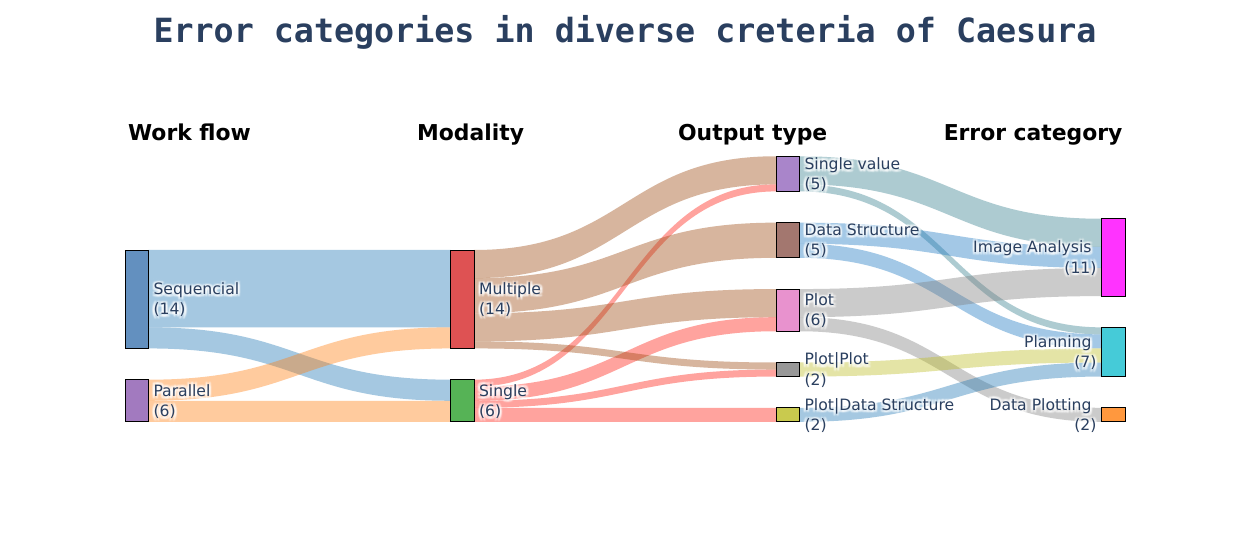}}
    \subfloat[M$^2$EX (ArtWork)]{\includegraphics[trim={25pt 25pt 25pt 32pt},clip, width=0.48\linewidth]{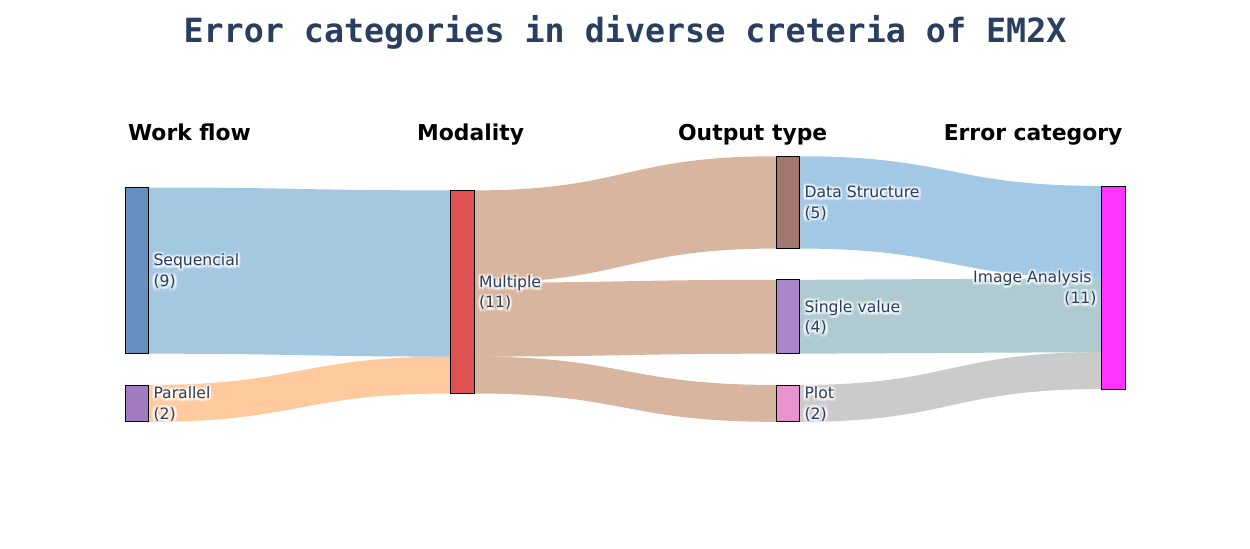}} \\
    \subfloat[CAESURA (RotoWire)]{\includegraphics[trim={25pt 25pt 25pt 32pt},clip, width=0.48\linewidth]{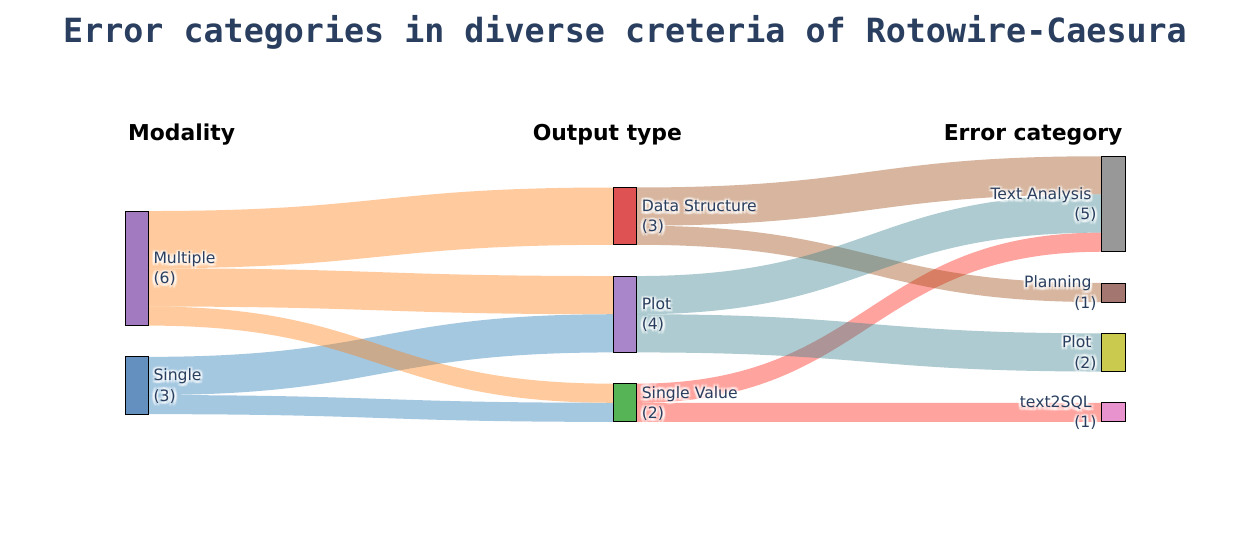}}
    \subfloat[M$^2$EX (RotoWire)]{\includegraphics[trim={25pt 25pt 25pt 32pt},clip, width=0.48\linewidth]{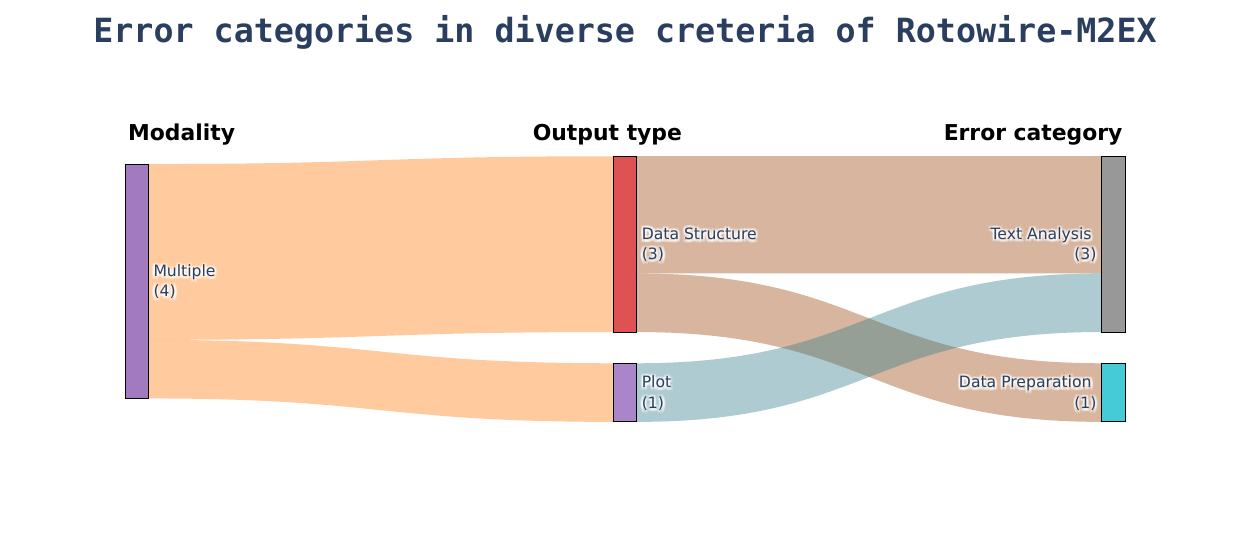}}\\
    \subfloat[M$^2$EX (EHRXQA)]{\includegraphics[trim={25pt 25pt 25pt 32pt},clip, width=0.48\linewidth]{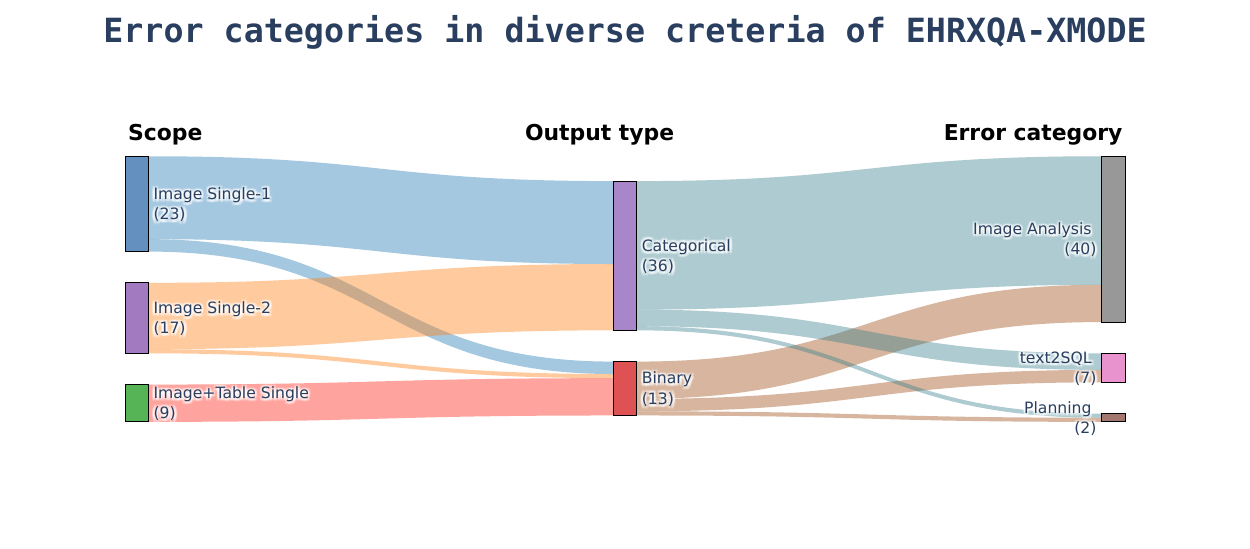}} \\
    \caption{Error analysis on different datasets: (a) CAESURA on ArtWork, (b) M$^2$EX on ArtWork, (c) CAESURA on RotoWire, (d) M$^2$EX on RotoWire, and (e) M$^2$EX on EHRXQA.}
    
    \label{fig:error-analysis}
\end{figure}
\paragraph{Error Analysis on the ArtWork Dataset}
As illustrated in Figure \ref{fig:error-analysis} (a), a total of 20 errors are identified out of 30 inference tasks for CAESURA. Of these, 14 errors occur within CAESURA’s sequential workflow. The errors include three single-modal questions and 11 multi-modal questions. Among the three single-modal, one task could not be resolved due to insufficient data available in the data pool. Following this failure, CAESURA attempts to replan twice but ultimately generates an incorrect plan, and consequently results in an erroneous response. The remaining two errors in single-modal tasks were classified as \textit{Plot Generation Errors}, which are caused by inconsistencies in the time axis units of the plot output.

For 11 errors in multi-modal questions, five are related to single-value outputs, four to plots, and three to data structures. All of these errors are attributed to incorrect outputs generated by the image analysis model. After further research, we found two ambiguous tasks in classifying the error categories. \textit{(1) Plot the number of paintings that depict war for each year} and \textit{(2) What is depicted on the oldest religious artwork in the database?} Both tasks failed due to improperly parsed sub question for the image analysis task, specifically the oversimplified term “war.” While this term is semantically related to the correct natural language question, “Does the image depict war?”, it does not fully capture the intent of the task. As a result, it cannot be classified as a completely faulty question. Notably, the M$^2$EX model generated correct results for these tasks, underscoring the limitations of CAESURA’s approach in handling subtle semantic distinctions.

In questions which require a parallel workflow - including two data structures,  plot | plot, and  plot | data structure outputs — errors are observed at the early planning stage. Our analysis reveals that CAESURA encounters significant challenges in generating accurate plans for embarrassingly parallel tasks. For two of these tasks, the system fails to generate any plan at all. For the remaining four tasks, CAESURA can provide partial results for some subtasks, but other subtasks are left unanswered, reflecting a broader issue in its ability to manage parallel planning.
Our M$^2$EX system successfully generates the appropriate plans for all tasks, as shown in Figure \ref{fig:error-analysis} (b). In addition, all text-to-SQL steps, data preparation pipelines, and plot outputs, where required, are validated as correct. As illustrated in Figure \ref{fig:error-analysis}(b), the only source of errors is the inaccurate output of the image analysis model, which accounted for 11 errors.
 No other errors are located in the text-to-SQL task, plot generation, or task planning deficiencies. 
This analysis highlights the image analysis model as the bottleneck in system performance, underscoring the need for further refinement in its predictive accuracy.

\paragraph{Error Analysis on the RotoWire Dataset}

Figure \ref{fig:error-analysis} (c) reveals that CAESURA encounters 9 errors across 12 inference tasks on the RotoWire dataset. These tasks are evenly divided between single-modal and multi-modal categories. Among the three single-modal tasks, one stumbles due to an SQL query missing essential filter clauses, resulting in inaccurate structured data. The other two, focused on plotting, fail to generate visualizations consistent with the analytical findings.

In the multi-modal group, six tasks face challenges. A task requiring a single-value output is derailed by suboptimal text analysis. Additionally, the Bart model’s limited text comprehension hampers two tasks expecting data structure outputs and two others involving plots, all undermined by faulty text interpretation. Another task, aimed at producing a structured output, falters during the planning stage because the strategy cannot be refined within the permitted attempts.

In contrast, our M$^2$EX system, as illustrated in Figure \ref{fig:error-analysis} (d), excels by devising suitable plans for all tasks and accurately resolving every single-modal task. However, it encounters issues in four multi-modal tasks: two demanding data structures and one plotting task succumb to flawed text analysis, while a fourth task needing a structured output fails during post-data preparation. Beyond these, no errors arise in text-to-SQL conversions or plot generation. This comparison underscores M$^2$EX’s greater resilience while highlighting text analysis as a shared weakness. CAESURA, however, suffers from additional pipeline limitations.

\paragraph{Error Analysis on the EHRXQA Dataset}

Since NeuralSQL is a one-step approach lacking task planning and explainability, we are unable to localize the source of errors as systematically as in the M$^2$EX or CAESURA systems. Consequently, we focus our error analysis solely on the M$^2$EX system using the EHRXQA dataset.

Figure \ref{fig:error-analysis} (e) presents the distribution of 49 errors across various steps, categorized by their respective scopes: \textit{Image Single-1} (23 errors), \textit{Image Single-2} (17 errors), and \textit{Image+Table Single} (9 errors). Among these, 36 errors are associated with the categorical scope, with 20 attributed to \textit{Image Single-1} and 16 to \textit{Image Single-2}. In contrast, errors linked to the binary output type are primarily found in the \textit{Image+Table Single} scope. Specifically, \textit{Image Single-1} contributes three binary errors, \textit{Image Single-2} accounts for one, and \textit{Image+Table Single} includes nine, summing up to 13 binary errors out of the total 49.
Considering the uneven distribution of errors across various output types and scopes, we identified inaccurate image analysis — primarily driven by the M3AE model \citep{10.1007/978-3-031-16443-9_65} — as the main source of errors. Our analysis reveals that errors linked to categorical output types (36) are nearly three times higher than those associated with binary output types (13). This suggests that the error pattern is less related to the task difficulty across different scopes and more influenced by the output type, as binary questions demonstrate a statistically higher success rate compared to categorical ones. Notably, the \textit{Image + Table Single} scope exclusively utilizes binary output types.

To gain a deeper understanding, a step-by-step error analysis reveals that out of the 23 errors in the \textit{Image Single-1} scope, 22 are due to inaccuracies in image analysis, while only one is related to a misstep in the text-to-SQL process. The specific question text for this case is: \textit{“Catalog all the anatomical findings seen in the image, given the first study of patient 11801290 on the first hospital visit.”} The generated SQL query fails to include the condition specifying the \textit{first study}, resulting in an incorrect output.
In the \textit{Image Single-2} category, 16 out of 17 total errors are due to inaccurate image analysis, with one error attributed to the text-to-SQL step. The specific query in question is: \textit{“Does the second-to-last study of patient 16345504 this year reveal still-present fluid overload/heart failure in the right lung compared to the first study this year?”}. The text-to-SQL task fails to correctly retrieve the \textit{first and last study of this year} as required, instead erroneously returning multiple studies from the current year.
In the \textit{Image+Table Single} scope, all nine errors involve binary output types. Of these, six result from inaccurate image analysis, one from incomplete planning, and two from an incorrect text-to-SQL step. The error caused by incomplete planning occurs with the question: \textit{“Did patient 19055351 undergo the combined right and left heart cardiac catheterization procedure within the same month after a chest x-ray revealed any anatomical findings until 2104?”}. In this case, the plan omits the necessary image analysis step, leading to an incorrect final output. During the reasoning stage, instances were identified where an empty output produced a \textit{no} response that coincidentally aligned with the ground truth. However, M$^2$EX’s explainability highlights this as a misclassification, as the absence of output was not due to correct reasoning.

Two errors in the \textit{Image+Table Single} category are attributed to text-to-SQL misbehavior. The specific questions causing these errors are: \textit{"Was patient 12724975 diagnosed with hypoxemia until 1 year ago, and did a chest x-ray reveal any tubes/lines in the abdomen during the same period?”} and \textit{"Was patient 10762986 diagnosed with a personal history of tobacco use within the same month after a chest x-ray showing any abnormalities in the aortic arch until 1 year ago?"} In both cases, the SQL queries fail to correctly apply the condition \textit{(since current time) until 1 year ago}, instead treating \textit{1 year ago} as a fixed point in time.

These findings highlight the pivotal role of accurate image analysis in multi-modal data exploration systems. Particularly, they emphasize a formidable challenge associated with categorical outputs. Moreover, the findings underscore the necessity of robust planning and effective SQL query generation to achieve optimal system performance. 
Addressing these challenges requires advancements in visual reasoning, temporal logic comprehension, and SQL generation, all of which are essential for mitigating errors and enhancing system accuracy.
%Use \verb|\appendix| before any appendix section to switch the section numbering over to letters. See Appendix~\ref{sec:appendix} for an example.

\end{document}